\documentclass[sigconf]{acmart}

\AtBeginDocument{%
  \providecommand\BibTeX{{%
    \normalfont B\kern-0.5em{\scshape i\kern-0.25em b}\kern-0.8em\TeX}}}

\setcopyright{acmcopyright}
\copyrightyear{2022}
\acmYear{2022}
\setcopyright{acmcopyright}\acmConference[KDD '22]{Proceedings of the 28th ACM
SIGKDD Conference on Knowledge Discovery and Data Mining}{August 14--18,
2022}{Washington, DC, USA}
\acmBooktitle{Proceedings of the 28th ACM SIGKDD Conference on Knowledge
Discovery and Data Mining (KDD '22), August 14--18, 2022, Washington, DC, USA}
\acmPrice{15.00}
\acmDOI{10.1145/3534678.3539369}
\acmISBN{978-1-4503-9385-0/22/08}

\usepackage{mathtools}

\usepackage{subfigure}
\usepackage{multirow}
\usepackage{bm}
\usepackage{bbm}
\usepackage{dsfont}
\usepackage{color,soul}
\usepackage{listings}
\usepackage{xcolor}
\usepackage{algorithmic}

\usepackage{natbib}

\usepackage[linesnumbered,vlined,ruled,noend]{algorithm2e}

\newcommand{\para}[1]{{\vspace{2pt} \bf \noindent #1 \hspace{1pt}}}
\definecolor{codegray}{rgb}{0.5,0.5,0.5}

\usepackage{enumitem}

\begin{document}

\title{Transfer Learning based Search Space Design for Hyperparameter Tuning}

\author{Yang Li$^{\dagger\mathsection}$, Yu Shen$^{\dagger}$, Huaijun Jiang$^\dagger$, 
Tianyi Bai$^*$, Wentao Zhang$^\dagger$, Ce Zhang$^\ddagger$, Bin Cui$^{\dagger\diamond}$}

\affiliation{
$^\dagger$
School of CS \& Key Laboratory of High Confidence Software Technologies (MOE), Peking University\country{China}
}

\affiliation{
$^\mathsection$Data Platform, TEG, Tencent Inc.\country{China}
}

\affiliation{
$^\ddagger$Department of Computer Science, Systems Group, ETH Z\"urich\country{Switzerland}
}

\affiliation{
$^*$School of Mathematics and Statistics, Beijing Institute of Technology\country{China}
}

\affiliation{
$^\diamond$Institute of Computational Social Science, Peking University (Qingdao)\country{China}
}

\affiliation{
$^\dagger$\{liyang.cs, shenyu, jianghuaijun, wentao.zhang, bin.cui\}@pku.edu.cn~~~~~
$^\ddagger$ce.zhang@inf.ethz.ch~~~~~$^\mathsection$baitianyi@bit.edu.cn\country{}
}\country{}

\renewcommand{\authors}{Yang Li, Yu Shen, Huaijun Jiang, Tianyi Bai, Wentao Zhang, Ce Zhang, and Bin Cui}
\renewcommand{\shortauthors}{Li et al.}

\begin{abstract}
The tuning of hyperparameters becomes increasingly important as machine learning (ML) models have been extensively applied in data mining applications.
Among various approaches, Bayesian optimization (BO) is a successful methodology to tune hyperparameters automatically. While traditional methods optimize each tuning task in isolation, there has been recent interest in speeding up BO by transferring knowledge across previous tasks. 
In this work, we introduce an automatic method to design the BO search space with the aid of tuning history from past tasks.
This simple yet effective approach can be used to endow many existing BO methods with transfer learning capabilities.
In addition, it enjoys the three advantages: universality, generality, and safeness.
The extensive experiments show that our approach considerably boosts BO by designing a promising and compact search space instead of using the entire space, and outperforms the state-of-the-arts on a wide range of benchmarks, including machine learning and deep learning tuning tasks, and neural architecture search.

\end{abstract}




\maketitle

\section{Introduction}
\label{sec:intro}

The performance of modern machine learning (ML) and data mining
methods highly depends on their hyperparameter configurations~\cite{he2016deep, hinton2012deep,goodfellow2016deep,devlin2018bert}, e.g., learning rate, the number of hidden layers in a deep neural network, etc.
As a result, automatically tuning the hyperparameters has attracted lots of interest from both academia and industry~\cite{quanming2018taking,bischl2021hyperparameter}.
A large number of approaches have been proposed to automate this process, including random search~\cite{bergstra2012random}, Bayesian optimization~\cite{hutter2011sequential,bergstra2011algorithms,snoek2012practical}, evolutionary optimization~\cite{hansen2016cma,real2019regularized} and bandit-based methods~\cite{jamieson2016non,li2018hyperband,falkner2018bohb,li2021mfes,li2022hyper}.

Among various alternatives, Bayesian optimization (BO) is one of the most prevailing frameworks for automatic hyperparameter optimization (HPO)~\cite{hutter2011sequential, bergstra2011algorithms, snoek2012practical,boreview}.
The main idea of BO is to use a surrogate model, typically a Gaussian Process (GP)~\cite{rasmussen2004gaussian}, to model the relationship between a hyperparameter configuration and its performance (e.g., validation error), and then utilize this surrogate to guide the configuration search over the given hyperparameter space in an iterative manner. 
However, with the rise of big data and deep learning (DL) techniques, the following two factors greatly hampers the efficiency of BO: (a) large search space and (b) computationally-expensive evaluation for each configuration.
Given a limited budget, BO methods can obtain only a few observations over the large space.
In this case, BO fails to converge to the optimal configuration quickly, which we refer to as the low-efficiency issue~\cite{falkner2018bohb,li2021volcanoml}.

\begin{figure*}[htb]
	\centering
		\scalebox{.9}[.9] {
		\includegraphics[width=1\linewidth]{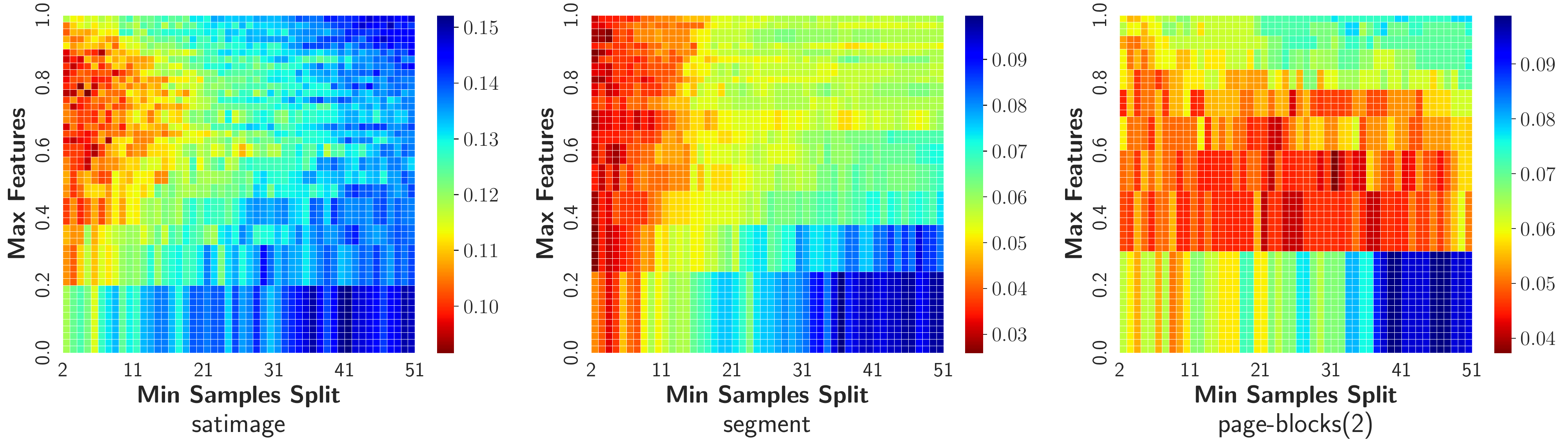}
         }
    \vspace{-1.em}
	\caption{Validation error of 2500 configurations (50 settings of max features and 50 settings of min samples split) when tuning two hyperparameters of random forest on three OpenML datasets. The red region indicates the promising configurations.}
    \vspace{-1.1em}
    \label{fig:intro_heatmap}
\end{figure*}

{\bf (Opportunities) }
To address this issue, researchers in the HPO community propose to incorporate the spirit of transfer learning to accelerate hyperparameter tuning, which could borrow strength from the past tasks (source tasks) to accelerate the current task (target task).
A line of work~\cite{feurer2018scalable,perrone2018scalable,wistuba2016two} aims to learn surrogates with the aid of past tuning history, and the surrogates are utilized to guide the search of configurations.
Orthogonal to these approaches, we concentrate on how to design a more promising and compact search space with the merit of transfer learning, and further speed up the HPO process.
The basic idea is that although the optimal configuration may be different for each HPO task, the region of well-performing hyperparameter configurations for the current task may share some similarity with previous HPO tasks due to the relevancy among tasks (See the left and middle heatmaps in Figure~\ref{fig:intro_heatmap}).
Motivated by the observation, in this paper, {\em we focus on developing an automatic search space design method for BO, instead of using the entire and large search space.}

{\bf (Challenges) } 
However, designing the compact search space automatically is non-trivial. To fully unleash the potential of transfer learning for space design, we need to consider the following problems: (a) {\em Representation of good/promising region}: The region of good configurations (promising region) in source tasks has uncertain shapes (See the deep red regions in Figure~\ref{fig:intro_heatmap}); in this case, simple geometrical representations of the search space (e.g., bounding box or ellipsoid used in ~\cite{pNIPS2019_9438}) cannot capture the shape of promising regions well. (b) {\em Relevancy between tasks}: While sharing some correlation between tasks, the underlying objective surface may be quite different (See the middle and right heatmaps in Figure~\ref{fig:intro_heatmap}). 
If one ignores this diversity, the performance of transfer learning may be greatly hampered, where a loose search space is obtained, or the optimal configuration is dismissed from the generated search space, which further leads to the ``negative transfer'' issue~\cite{pan2010survey}.
Several HPO methods~\cite{feurer2015initializing,wistuba2016two} measure the similarity scores between HPO tasks by computing the distance of meta-features for data set, while the meta-features in practice are often hard to obtain and need careful manual design~\cite{feurer2018scalable,pNIPS2019_9438}.
Therefore, we need to learn this similarity during the HPO process automatically.

In addition, (c) {\em utilization of promising regions} is also a challenging problem. 
To prevent negative transfer, the dissimilar tasks should hold large enough promising regions, so that the optimal configuration of the target task is not excluded.
On the contrary, given similar source tasks, a small promising region could greatly speed up the search of configuration in the target task.
As a result, the size of promising regions should be carefully crafted based on task relevancy.
Finally, we need to leverage these promising regions to build a compact search space for the target task.
Rather than simple intuitions that use a single promising region, a mature design should utilize multiple source tasks.
However, how to select and leverage those promising regions is still an open question.

In this paper, we propose a novel transfer learning-based search space design method for hyperparameter optimization.
Instead of restricting the promising region to geometrical shapes, we propose to use a machine learning algorithm to learn the good region automatically. Concretely, we turn it into a supervised classification problem. 
In addition, we develop a ranking-based method to measure the task correlation between the source and target task on the fly, and the size of promising regions could be adjusted adaptively based on this correlation.
With these basic ingredients, we first extract the promising region from each source task based on the task similarity, respectively, and then generate the final search space for the target task -- a sub-region of the complete search space -- via sampling-based framework along with a voting mechanism.
In this way, the proposed method could leverage the promising regions obtained from similar previous HPO tasks to craft a promising and compact search space for the current task automatically. 

{\bf (Contributions) } 
We summarize our main contributions as follows:
(a) We present a novel transfer learning-based search space design method for hyperparameter optimization. Our method learns a suitable search space in an adaptive manner: it can endow BO algorithms with transfer learning capabilities.
(b) On an extensive set of benchmarks, including tuning ML algorithm on OpenML problems, optimizing ResNet on three vision tasks, and conducting neural architecture search (NAS) on NASBench201, the empirical results demonstrate that our approach outperforms the existing method and significantly speeds up the HPO process.
(c) Despite the empirical performance, our method enjoys the following three advantages: universality, practicality, and safeness. With these properties, our approach can be seamlessly combined with a wide range of existing HPO techniques.
(d) We create and publish large-scale benchmarks for transfer learning of HPO, which are significantly larger than the existing ones. We hope this benchmark would help facilitate the research on search space design for HPO.

\section{Related Work}

Hyperparameter optimization (HPO) is one of the most fundamental tasks when developing machine learning (ML) and data mining (DM) applications~\cite{10.1145/3447548.3470827,bischl2021hyperparameter}. 
Many approaches have been proposed to conduct automatic HPO, such as random search~\cite{bergstra2012random}, Bayesian optimization (BO)~\cite{hutter2011sequential,bergstra2011algorithms}, bandit based method~\cite{jamieson2016non,li2018hyperband,li2020efficient}, etc.
In addition, many tuning systems~\cite{openbox,Amazon_SageMaker,oboe,autotune,li2022hyper} in the ML/DM community have successfully integrate the HPO algorithms.

Recently, many researchers have proposed to utilize transfer learning within the BO framework to accelerate HPO.
The target is to leverage auxiliary knowledge acquired from the previous source tasks to achieve faster optimization on the target task, that is, borrow strength from past HPO tasks.
Orthogonal to our contribution, one common way is to learn surrogate models from past tuning history and use them to guide the search of hyperparameters.
For instance, several methods learn all available information from both source and target tasks in a single surrogate, and make the data comparable through multi-task GPs~\cite{swersky2013multi}, a ranking algorithm~\cite{bardenet2013collaborative}, a mixed kernel GP~\cite{yogatama2014efficient}, the GP noisy model~\cite{joy2016flexible}, a multi-layer perceptron with Bayesian linear regression heads~\cite{snoek2015scalable,perrone2018scalable}, a Gaussian copula process~\cite{salinas2020quantile} or replace GP with Bayesian neural networks~\cite{springenberg2016bayesian}.
In addition, several approaches train multiple base surrogates, and then combine all base surrogates into a single surrogate with dataset similarities~\cite{wistuba2016two}, weights adjusted via GP uncertainties~\cite{schilling2016scalable} or the weights estimated by rankings~\cite{feurer2018scalable}. 
Similarly, ~\citet{golovin2017google} build a stack of GPs by iteratively regressing the residuals with the most recent source task.
Finally, instead of fitting surrogates on the past observations, several approaches~\cite{DBLP:journals/corr/abs-1709-04636,kim2017learning} achieve transfer learning in a different way. 
They warm-start BO by selecting several initial hyperparameter configurations as the start points of search procedures to accelerate the searching process.

Recently, transferring search space has become another way for applying transfer learning in HPO. ~\citet{wistuba2015hyperparameter} prune the bad regions of search space according to the results from previous tasks. 
This method suffers from the complexity of obtaining meta-features and relies on some other parameters to construct a GP model. 
On that basis, ~\citet{pNIPS2019_9438} propose to utilize previous tasks to design a sub-region of the entire search space for the new task. 
However, this method ignores the similarity between HPO tasks, and applies a simple low-volume geometrical shape (bounding box or ellipsoid) to obtain the sub-region that contains the optimal configurations from all past tasks. 
Therefore, it may design a loose search space or exclude the best configurations and further lead to negative transfer~\cite{pan2010survey}.
This work is the most closely related to our method. 
We want to highlight that the other forms of transfer (e.g., surrogate-based transfer or warm-starting BO) are orthogonal to the search space transfer methods, and thus these approaches can be seamlessly combined to pursue better performance of HPO.

\section{Preliminary}
In this section, we first introduce the problem definition and then describe the basic framework of Bayesian optimization.

\subsection{HPO over a Reduced Search Space}
\label{sec:reduced}
The HPO of ML algorithms can be modeled as a black-box optimization problem. 
Given a common hyperparameter space $\mathcal{X}$ and tuning history from previous HPO tasks, we need to optimize the current tuning task.
The goal of this HPO problem is to find the best hyperparameter configuration that minimizes the objective function $f^T$ on the target task, which is formulated as follows,
\begin{equation}
    \mathop{\arg\min}_{\bm{x} \in \mathcal{X}}f^T(\bm{x}),
    \label{eq:original}
\end{equation}
where $f^T(\bm{x})$ is the ML model's performance metric (e.g., validation error) corresponding to the configuration $\bm{x}$. 
Due to the intrinsic randomness of most ML algorithms, we evaluate a configuration $\bm{x}$ and can only get its noisy observation $y = f(\bm{x}) + \epsilon$ with $\epsilon \sim \mathcal{N}(0, \sigma^2)$. 
In this work, we consider methods that output a compact search space $\hat{\mathcal{X}}\subseteq\mathcal{X}$ with the aid of tuning history from past tasks. 
Instead of Equation~\ref{eq:original}, we solve the following problem:
\begin{equation}
    \mathop{\arg\min}_{\bm{x} \in \hat{\mathcal{X}}}f^T(\bm{x}).
    \label{eq:new}
\end{equation}

While $\hat{\mathcal{X}}$ is much smaller than $\mathcal{X}$, optimization methods may find those optimal configurations faster. 
Therefore, we aim to design $\hat{\mathcal{X}}$ such that it contains a proper set of promising configurations, which is close to the good region in the original space $\mathcal{X}$.
 
\subsection{Bayesian Optimization}
\label{sec:bo}
Bayesian optimization (BO) works as follows. 
Since evaluating the objective function $f$ for a
given configuration $\bm{x}$ is expensive, it approximates $f$ using a surrogate model $M:p(f|D)$ fitted on observations $D$, and this surrogate is much cheaper to evaluate.
Given a configuration $\bm{x}$, the surrogate model $M$ outputs the posterior predictive distribution at $\bm{x}$, that is,
$f(\bm{x}) \sim \mathcal{N}(\mu_{M}(\bm{x}), \sigma^2_{M}(\bm{x}))$.
BO methods iterate the following three steps: 1) use surrogate $M$ to select a promising configuration $\bm{x}_n$ that maximizes the acquisition function $\bm{x}_{n}=\arg\max_{\bm{x} \in \mathcal{X}}a(\bm{x}; M)$, where the acquisition function is to balance the exploration and exploitation trade-off; 2) evaluate this point to get its performance $y_n$, and add the new observation $(\bm{x}_{n}, y_{n})$ to $D=\{(\bm{x}_j,y_j))\}_{j=1}^{n-1}$; 3) refit $M$ on the augmented $D$. 

Expected Improvement (EI) \cite{jones1998efficient} is a common acquisition function, and it is widely used in the HPO community for its excellent empirical performance. 
EI is defined as follows:
\begin{equation}
\label{eq_ei}
a(\bm{x}; M)=\int_{-\infty}^{\infty} \max(y^{\ast}-y, 0)p_{M}(y|\bm{x})dy,
\end{equation}
where $M$ is the surrogate model and $y^{\ast}$ is the best performance observed in $D$, i.e., $y^{\ast}=\min\{y_1, ..., y_n\}$. 
By maximizing this EI function $a(\bm{x}; M)$ over the hyperparameter space $\mathcal{X}$, BO methods can find a configuration with the largest EI value to evaluate for each iteration.
Algorithm~\ref{algo:bo} displays the BO framework. 
While the \textit{design} function does nothing in vanilla BO and returns the original space (Line 8), we aim to design a compact and promising search space to accelerate HPO as introduced in Section~\ref{sec:reduced}.

\begin{algorithm}[tb]
  \small
  \caption{Pseudo code for Bayesian Optimization}
  \label{algo:bo}
  \begin{algorithmic}[1]
  \REQUIRE the number of trials $T$, the hyper-parameter space $X$,  surrogate model $M$, acquisition function $\alpha$, and initial hyper-parameter configurations $X_{init}$.
  \FOR{\{$\bm{x} \in X_{init}\}$}
    \STATE evaluate the configuration $\bm{x}$ and obtain its performance $y$.
    \STATE augment $D = D \cup (\bm{x}, y)$.
  \ENDFOR
  \STATE initialize observations $D$ with initial design.
  \FOR{\{ $i = |X_{init}| + 1, ..., T\}$}
  \STATE fit surrogate $M$ based on observations $D$.
  \STATE design the search space: $\hat{\mathcal{X}}=\operatorname{design}(\mathcal{X})$.
  \STATE select the configuration to evaluate: $\bm{x}_i=\operatorname{argmax}_{\bm{x}\in \hat{\mathcal{X}}}\alpha(\bm{x}, M)$.
  \STATE evaluate the configuration $\bm{x}_i$ and obtain its performance $y_i$.
  \STATE augment $D=D\cup(\bm{x}_i,y_i)$.
  \ENDFOR
  \STATE \textbf{return} the configuration with the best observed performance.
\end{algorithmic}
\end{algorithm}

\section{The Proposed Method}
\label{sec4}

In this section, we introduce a novel search space design method for hyperparameter tuning. We first present the notations and overview of our method, then introduce the two critical steps: promising region extraction and target search space generation. Finally, we end this section with an algorithm summary and discussion.

\subsection{Notations and Overview}
Our method takes the observations from $K+1$ tasks as input, in which $D^1$, ..., $D^K$ are tuning history from $K$ source tasks and $D^T$ is the observations in the target task. 
The $i$-th source task contains $n^i$ evaluated configurations $D^i=\{(\bm{x}_j^i, y_j^i)\}_{j=1}^{n^i}$.
Unlike $\{D^i\}_{i=1}^K$ that are obtained in previous tuning procedures, the number of observations in $D^T$ grows along with the current tuning process.
After finishing $t$ trials, the target observations are $D^T=\{(\bm{x}_j^T, y_j^T)\}_{j=1}^{t}$.

In this work, we consider methods that take the previous observations $\{D^i\}_{i=1}^K$, and the current observations $D^T$ as inputs, and output a compact search space $\hat{\mathcal{X}}\subseteq\mathcal{X}$. 
Our proposed method designs the compact search space based on the promising regions obtained from source tasks. 
The \textbf{promising region} refers to a sub-region of the original search space $\mathcal{X}^i \in \mathcal{X}$ where the optimal configurations of the target task $\bm{x}^*$ are located with a high probability. 
Before optimizing the target task, our approach trains a surrogate model $M^i$ on observations $D^i$ from each source task $i$. 
These surrogate models can be fitted in advance.
To utilize the promising regions and address the challenges in Section~\ref{sec:intro}, our search space design method includes two critical steps: (1) promising region extraction and (2) target search space generation. 
In short, our approach applies the machine learning algorithm to represent complex promising regions and adopt a ranking-based method to measure the task correlation.
Then, it generates the target search space via a sampling framework along with the voting mechanism. 
We will introduce the details in Sections~\ref{sec:prg} and~\ref{sec:prc}, respectively.

\subsection{Promising Region Extraction}
\label{sec:prg}
To transfer the search spaces, our method first extracts a promising region $\mathcal{X}^i$ from each source task $i$. 
A simple intuition is that when the target task is quite similar to a source task, the region of the best configurations are similar in both two tasks, 
so we have high confidence to extract a relatively small region that the optimal configuration of the target task locates in that region.
On the contrary, when the target task is quite different from a source task, we are not sure whether the best configurations of the source task work well on the target task,
so we need a conservatively large enough space to contain a sufficient number of candidate configurations.
Therefore, the size of the promising region should be carefully crafted.
In the following, we will answer two questions: 1) how to define the similarity between the source tasks and the target task, and 2) how to extract the promising region with a proper size according to the similarity.

We measure the task similarity between source tasks and the target task on the target observations $D^T$ via a ranking-based method.
In HPO, the ranking is more reasonable than the actual performance value of each configuration, and we care about the partial orderings over configurations, i.e., the region of promising configurations.
Therefore, we apply the ratio of order-preserving pairs to measure this similarity. The definition of the similarity $S(M^i;D^T)$ between the $i$-th source task and the target task is given by,
\begin{equation}
\small
\begin{aligned}
    F(M^i;D^T)& =\sum_{j=1}^{\left|D^T\right|} \sum_{k=j+1}^{\left|D^T\right|} \mathds{1}\left(\left(M^i(\bm{x}_j)<M^i(\bm{x}_k)\right)\otimes \left(y_j<y_k\right) \right) \\
    S(M^i;D^T)& =2F(M^i;D^T)/(|D^T|*(|D^T|-1)),
\end{aligned}
\label{eq:similarity}
\end{equation}
where $F(M^i;D^T)$ is the number of order-preserving pairs, $|D^T|$ is the number of target observations, $M^i(x_j)$ is the predictive mean of the surrogate $M^i$ of the $i$-th source task given the configuration $x_j$, and $\otimes$ is the exclusive-nor operation, in which the statement value is true only if the two sub-statements return the same value.

While previous work~\cite{pNIPS2019_9438} represents the promising region of a source task as a single area, we argue that the promising regions in many HPO tasks are non-convex and discontinuous, i.e., there may be several local optima, and the final promising region often consists of several good areas.
In addition, this method restricts the region with some geometrical shape, while the shape of promising regions in most real-world HPO tasks is uncertain and cannot be recognized easily.
To represent multiple promising areas along with uncertain shapes, We turn this problem into a binary classification task and employ the Gaussian Process Classifier (GPC) to learn the promising region, where the classifier could predict whether a configuration from the search space belongs to the promising region or not. 
For each source task $i$, our method prepares the training data $\{\bm{x^i_j},b^i_j\}^{|D^i|}_{j=1}$ as follows,
\begin{equation}
    b^i_j=\left\{
    \begin{array}{lcr}
     1    &    &  \text{if} \ y^i_j < y^i_+\\
     0    &    &  \text{if} \ y^i_j \geq y^i_+
    \end{array}\right.,
\label{eq:training_data}
\end{equation}
where $y^i_+$ is determined by some quantile $\alpha^i$ of performance values in $D^i$, so that the cumulative distribution function $P(y<y^i_+)=\alpha^i$. 
Inspired by the aforementioned intuition, we further propose to control the size of promising regions by adjusting $\alpha^i$ based on task similarity, and the adjustment rule is given by,
\begin{equation}
\small
    \alpha^i=\alpha_{min}+(1-2*\mathop{\max}(S(M^i;D^T)-0.5,0))*(\alpha_{max}-\alpha_{min}),
\label{eq:percentile}
\end{equation}
where $\alpha_{max}$ and $\alpha_{min}$ are two parameters set close to 1 and 0 respectively, which control the aggressiveness of the quantile $\alpha^i$. 
The two parameters ensure that in the worst case, when the source task is adverse to the target task, the size of the promising region will be determined by $\alpha_{max}$, i.e., the original complete search space.
When a source task is quite similar to the target task, the value of $\alpha^i$ approaches $\alpha_{min}$ so that a compact promising region can be extracted.
Finally, our method fits a GPC model $G^i$ on the training data in Eq.~\ref{eq:training_data} for each source task and extracts the promising region as $\mathcal{X}^i=\{\bm{x}_j|\bm{x}_j \in \mathcal{X},G^i(\bm{x}_j)=1\}$.
Note that, due to the growing observation $D^T$ during the HPO process, the promising region from each source task adaptively adjusts based on the task similarity.

\subsection{Target Search Space Generation}
\label{sec:prc}
While each source task holds a promising region, the next step is to combine those regions and generate the search space for the current iteration. 
An intuitive way is to select the most similar task and directly apply its promising region as the target space, that is:
\begin{equation}
    \hat{\mathcal{X}}=\mathcal{X}^m,\quad m=\mathop{\arg\max}_{i=1,...,k}S(M^i;D^T).
\end{equation}

However, since the search process could easily be trapped in a sub-optimal local region provided by the most similar source task, this design may lead to the over-exploitation issue over the whole search space. 
Another alternative is to sample a task according to the similarity and then apply its promising region, that is:
\begin{equation}
    \hat{\mathcal{X}}=\mathcal{X}^m, \quad m \sim \mathbb{P}(.),
\label{eq:sample}
\end{equation}
where $\mathbb{P}(.)$ is the distribution computed based on the similarity, in which $p(i)=S(M^i;D^T)/ \sum_{i=1}^K S(M^i;D^T)$.
Though sampling enables exploration on different promising regions, the source information is not fully utilized in optimization, i.e., the information from only one source task is used in this design of space. 

To encourage exploration and utilize more source tasks, our method adopts a sampling framework along with the voting mechanism.
It first samples $k$ source tasks out of $K$ tasks without replacement. 
Concretely, we scale the sum of similarity to 1 and sample source tasks based on this distribution.
We denote the sampled tasks as $s_1,...,s_k$.
Then, for each configuration $\bm{x}\in\mathcal{X}$, our method builds a voting ensemble $\hat{G}$ of $k$ GPC models as,
\begin{equation}
    \hat{G}(\bm{x}_j)=\left\{
    \begin{array}{lcl}
     1    &    &  \text{if} \ \sum_{i=1}^k G^{s_i}(\bm{x}_j) \geq \lfloor \frac{k}{2} \rfloor\\
     0    &    &  \text{else}
    \end{array}\right..
\label{eq:voting}
\end{equation}

When a configuration is in the majority of promising regions of the sampled tasks, it is regarded as a feasible configuration in the compact search space. 
The final target space is formulated as $\hat{\mathcal{X}}=\{\bm{x}_j|\bm{x}_j \in \mathcal{X},\hat{G}(\bm{x}_j)=1\}$.

\subsection{Algorithm Summary}
Algorithm \ref{algo:framework} gives the pseudo code of the \textit{design} function. 
The function is called during each iteration as shown in Algorithm~\ref{algo:bo}.
We first extract a promising region for each source task by computing the similarity based on Equation~\ref{eq:similarity} (Line 1) and training the GPC model based on the generated training data given by Equation~\ref{eq:training_data} (Line 2).
Then, we sample source tasks based on the similarity distribution (Line 3) and generate the target search space by combining those promising regions via the voting results of GPC models (Line 4).

\begin{algorithm}[tb]
  \small
  \caption{Pseudo code for \textit{design} function.}
  \label{algo:framework}
  \begin{algorithmic}[1]
  \REQUIRE the surrogates in $K$ source tasks: $M^{i}$ with $i\in [1, K]$, the target observation $D^T$, the configuration space $\mathcal{X}$.
  \STATE compute the similarity $S(M^i;D^T)$ between each source task $i$ and \newline the target task based on Eq.~\ref{eq:similarity}.
  \STATE Train a Gaussian Process Classifier $G^i$ for each source task $i$ based \newline on training data from Eq.~\ref{eq:training_data}.
  \STATE Sample $k$ source tasks $s_1,...,s_k$ based on similarity distribution \newline from Eq.~\ref{eq:sample}.
  \STATE Generate the target search space $\hat{\mathcal{X}}$ based on the voting results of \newline $G^{s_i}$ with $i\in [1,k]$.
  \STATE \textbf{return} the final search space $\hat{\mathcal{X}}$.
\end{algorithmic}
\end{algorithm}

\subsection{Discussion}
To our knowledge, our method is the first method that owns the following desirable properties simultaneously in the field of search space design.
\textbf{1. Universality.}
Our method focuses on accelerating HPO by designing promising and compact search space on the target task, so it is orthogonal to and compatible with other forms of transfer methods and a wide range of BO methods. In other words, it can be easily combined into those methods by integrating the \textit{design} function as in Algorithm~\ref{algo:bo}.
\textbf{2. Practicality.} 
A practical search space design method should support different types of hyperparameters, including both numerical and categorical ones. 
While previous work~\cite{pNIPS2019_9438} only works on numerical hyperparameters due to the use of geometrical shapes, our method employs an ML model to learn promising regions with uncertain shapes, so that it supports both two types of hyperparameters.
In addition, the computational complexity of \textit{design} function is $O(Kn^3)$, where $K$ is the number of source tasks and $n$ is the number of observations in each source task; due to the linear growth on $K$, the proposed method could scale to a large number of source tasks $K$ (good scalability).
\textbf{3. Safeness.}
The design of our method ensures that dissimilar source tasks may influence little to the final search space. 
Those tasks have little chance to be sampled due to their low similarity.
And once they are sampled by coincidence, Eq.~\ref{eq:percentile} returns a large percentile, so that their promising regions will be relatively large, and the optimal configuration of the target task is less likely to be excluded.

\section{Experiments and Results}
\label{exp_sec}

In this section, we evaluate the superiority of our approach from two perspectives: 1) the empirical performance compared with the existing search space design method on a wide range of benchmarks, 2) its advantages in terms of universality, practicality and safeness.

\begin{figure*}[htb]
	\centering
	\subfigure[winequality\_white]{
		\scalebox{0.23}[0.23]{
			\includegraphics[width=1\linewidth]{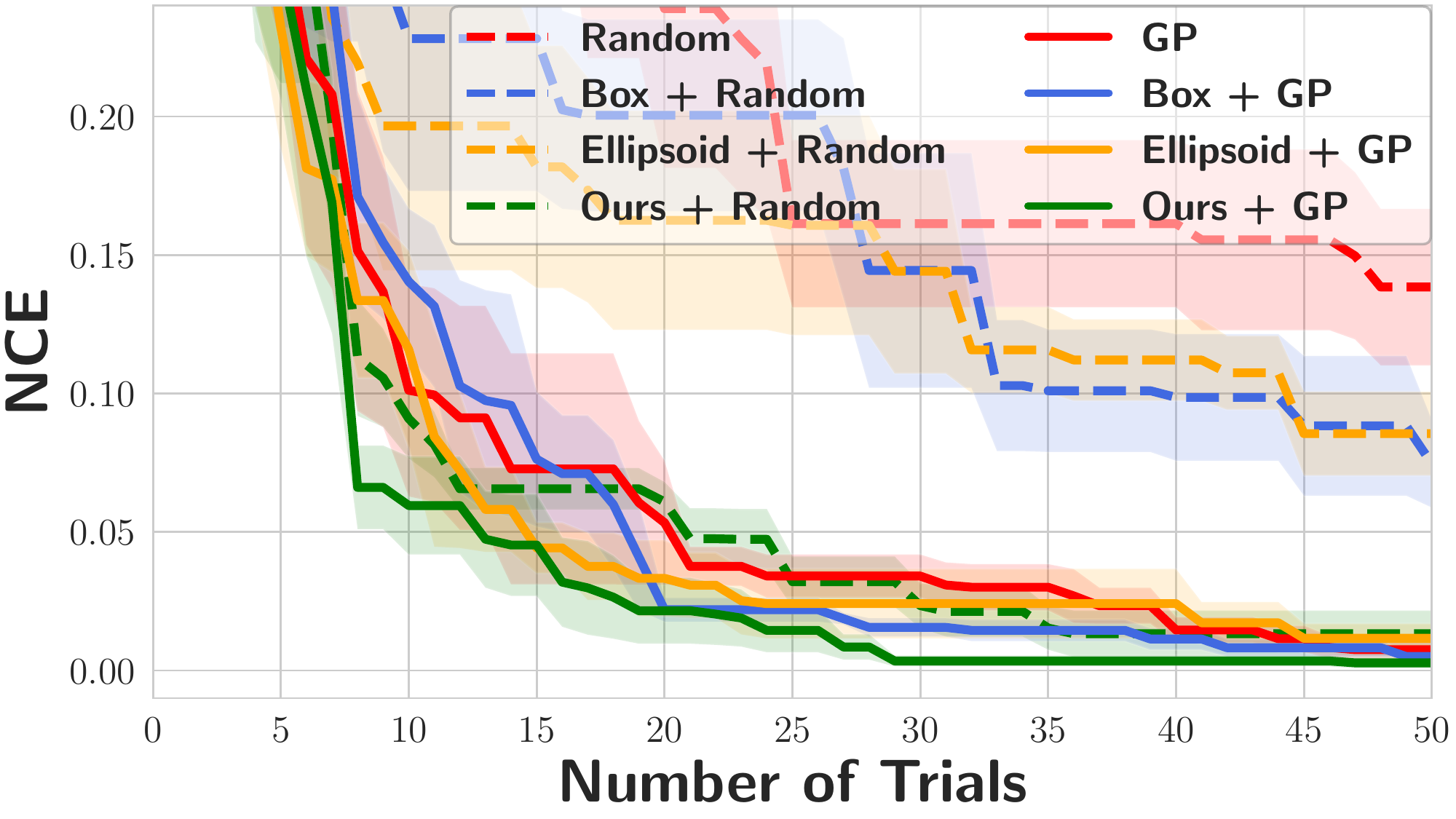}
	}}
	\subfigure[page-blocks(2)]{
		\scalebox{0.23}[0.23]{
			\includegraphics[width=1\linewidth]{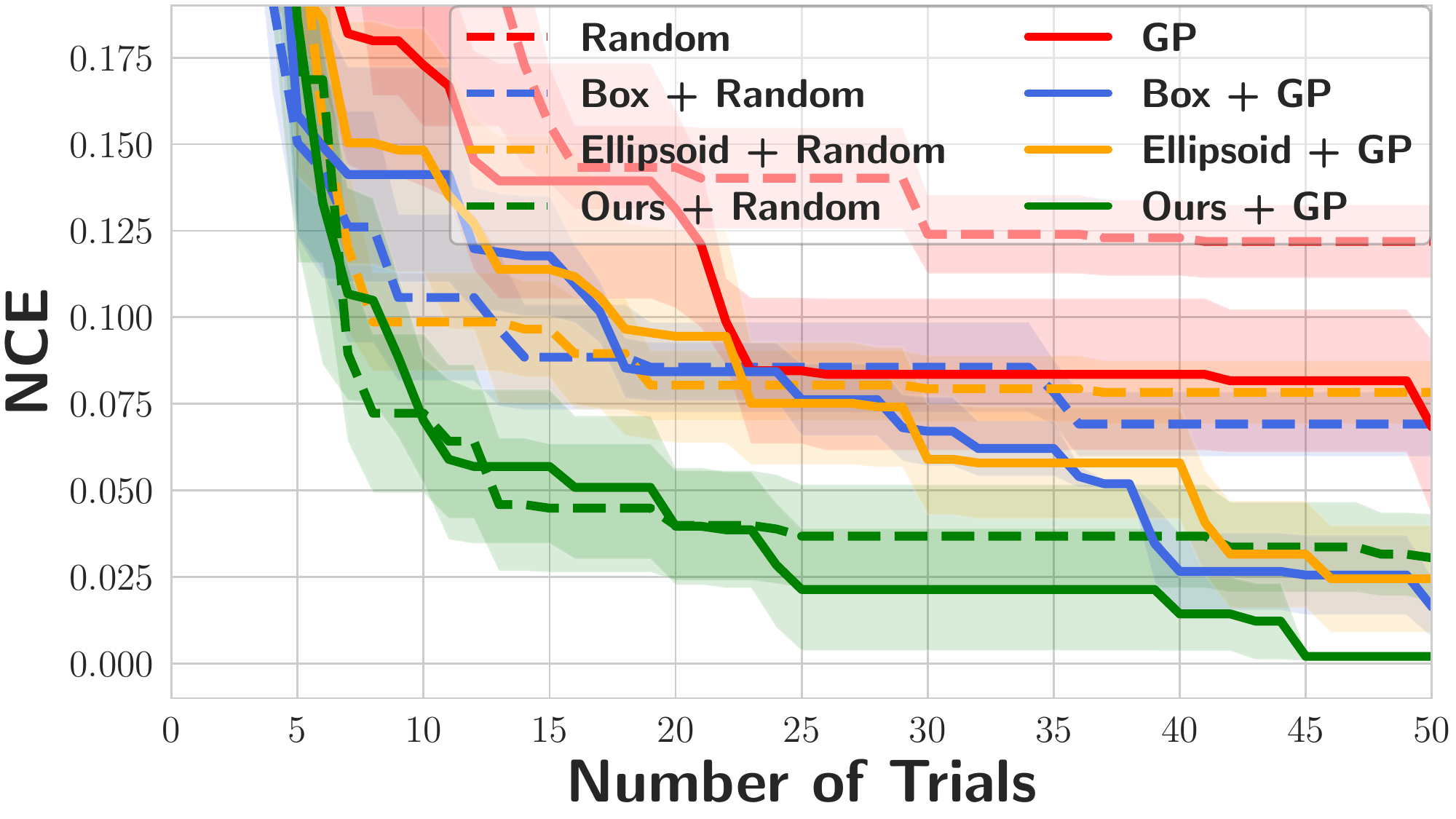}
	}}
	\subfigure[sick]{
		\scalebox{0.23}[0.23]{
			\includegraphics[width=1\linewidth]{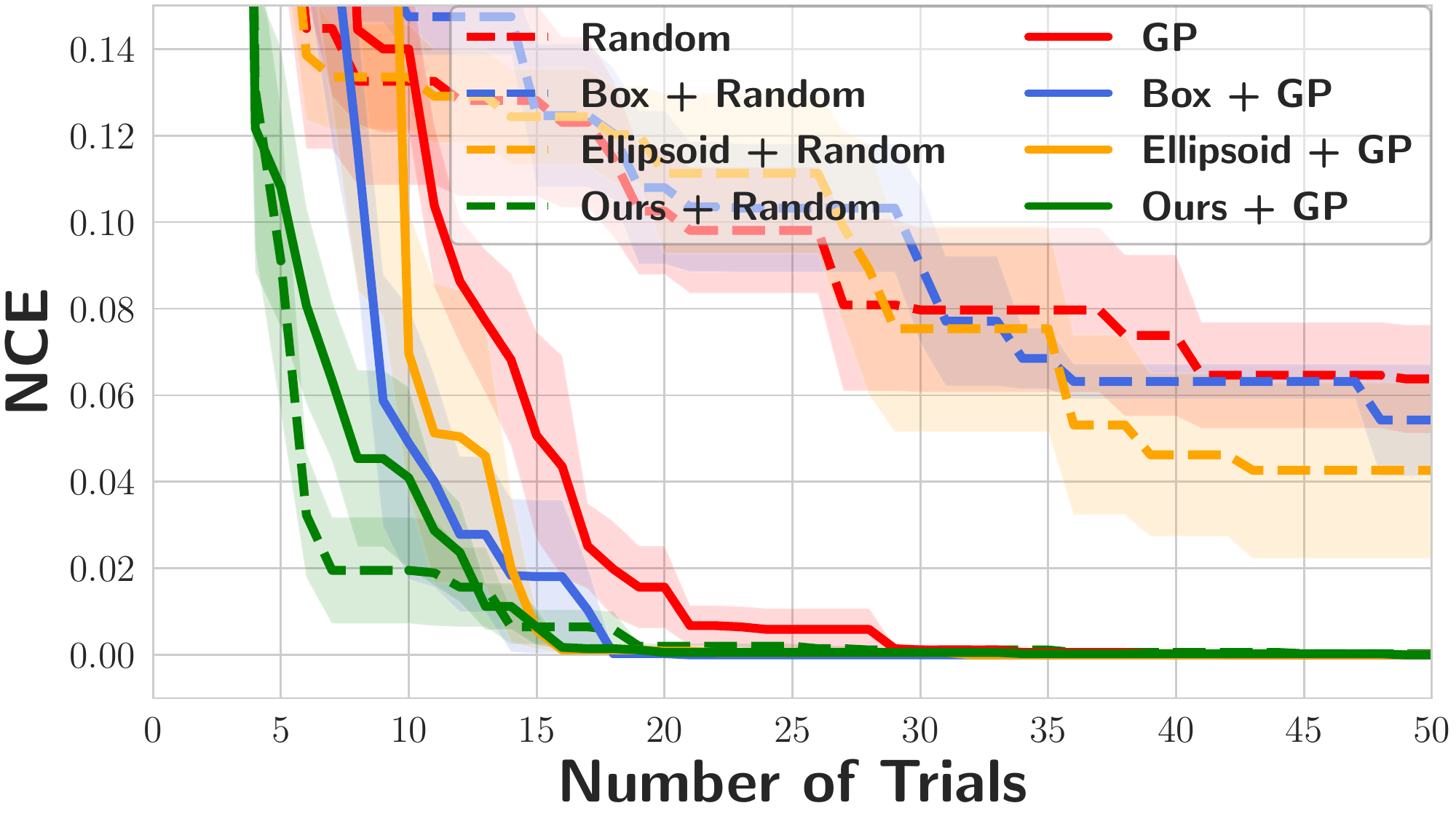}
	}}
    \subfigure[musk]{
		\scalebox{0.23}[0.23]{
			\includegraphics[width=1\linewidth]{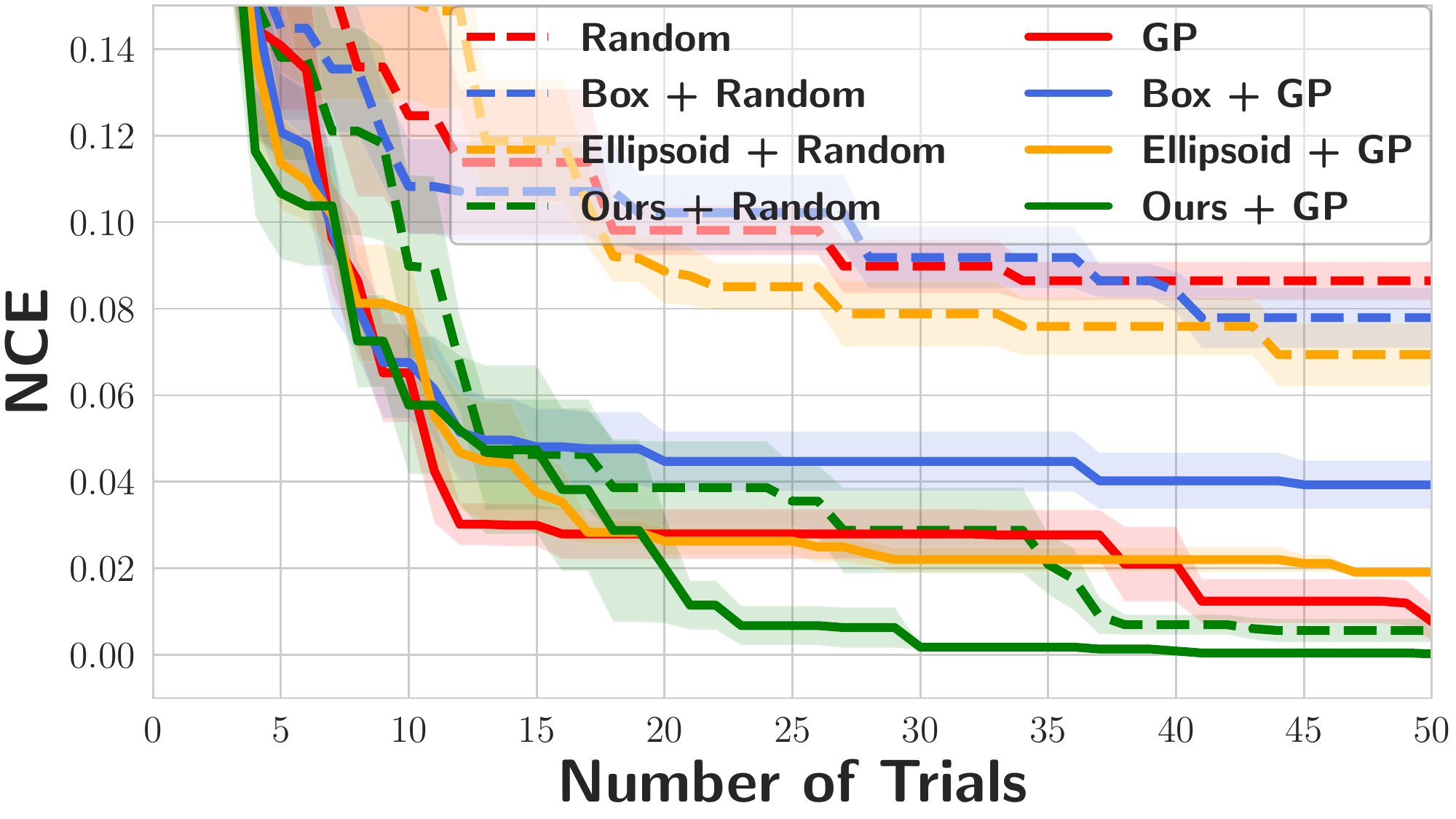}
	}}
	\quad
	\subfigure[puma8NH]{
		\scalebox{0.23}[0.23]{
			\includegraphics[width=1\linewidth]{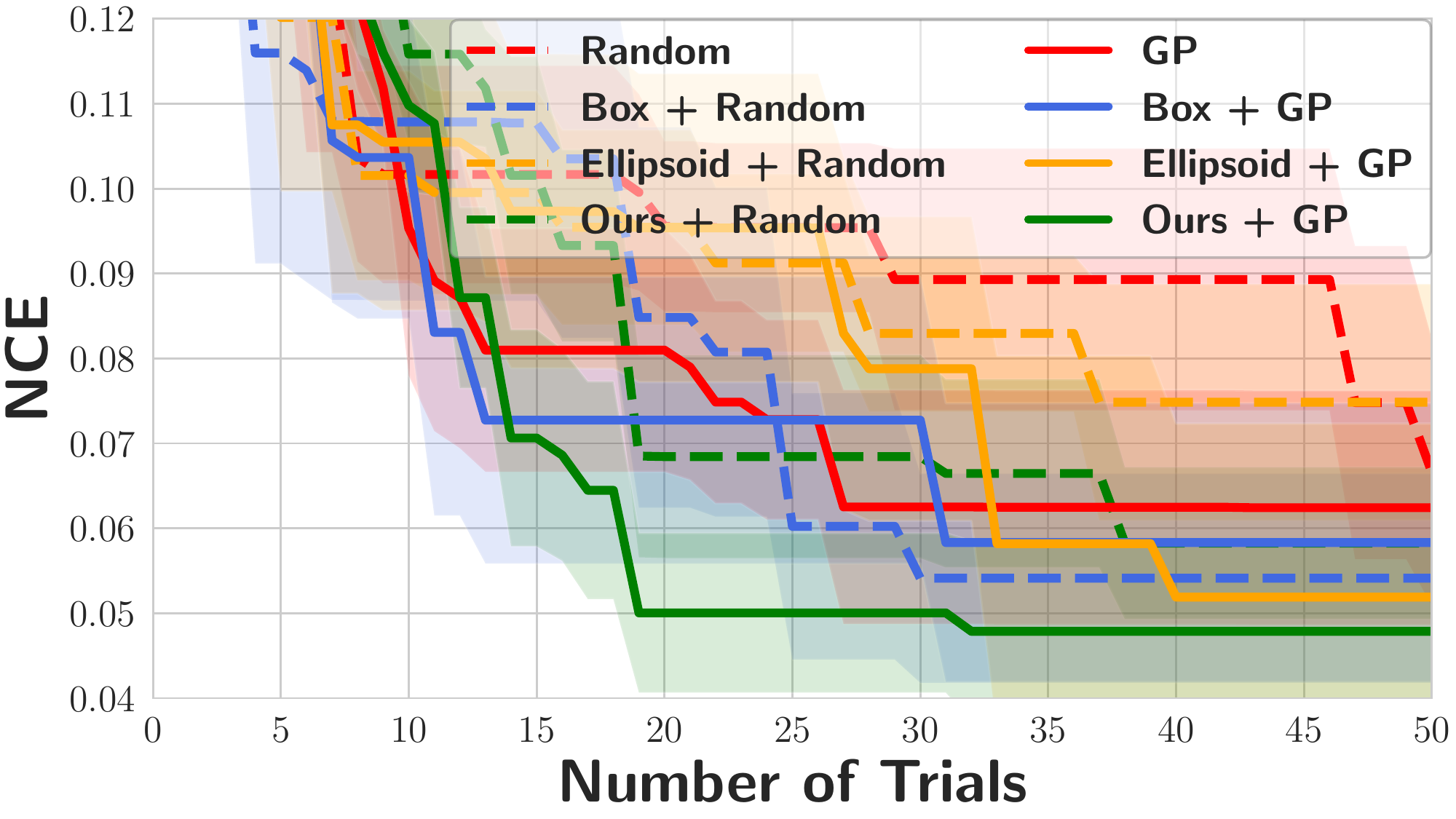}
	}}
	\subfigure[satimage]{
		\scalebox{0.23}[0.23]{
			\includegraphics[width=1\linewidth]{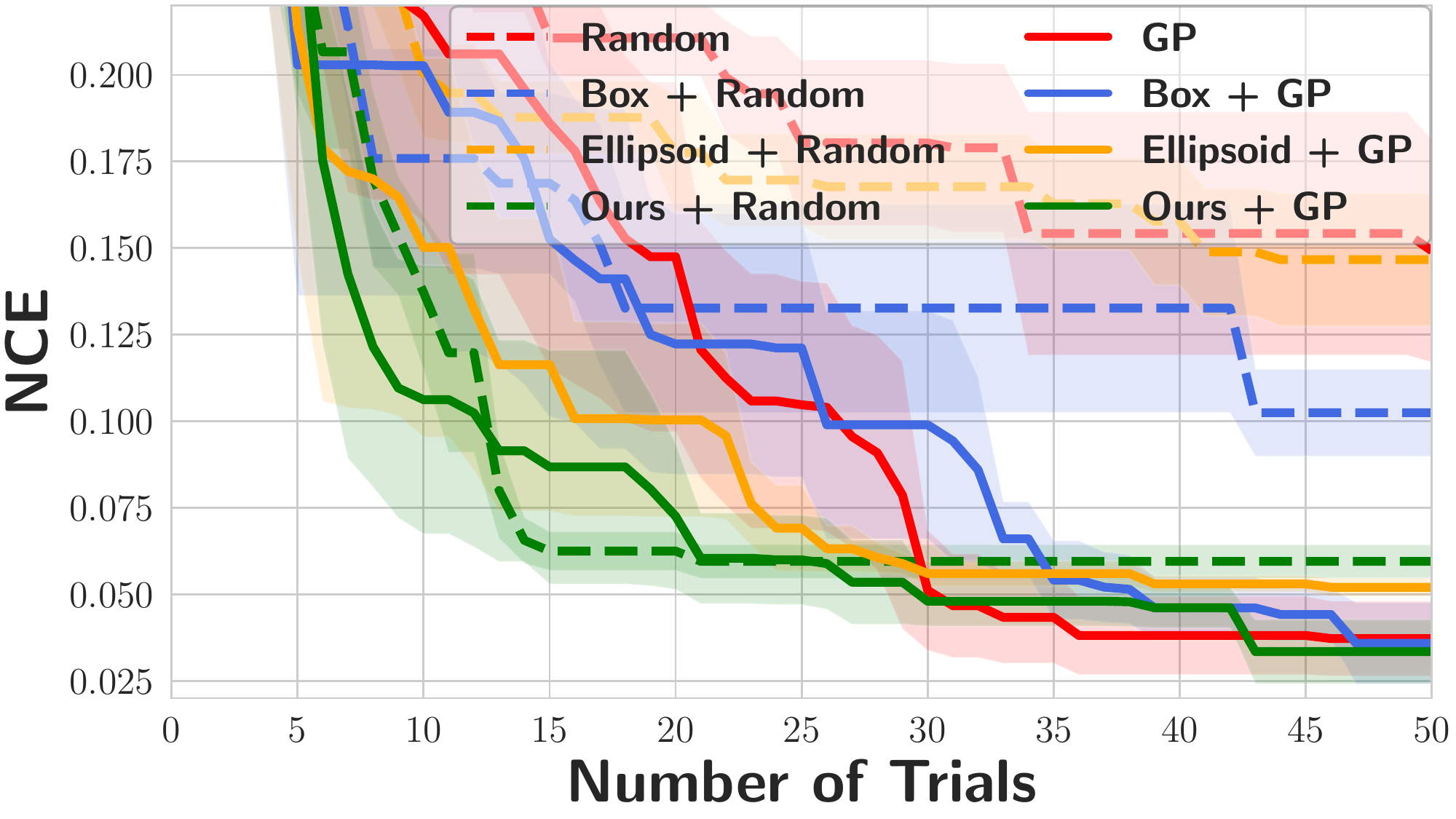}
	}}
	\subfigure[segment]{
		\scalebox{0.23}[0.23]{
			\includegraphics[width=1\linewidth]{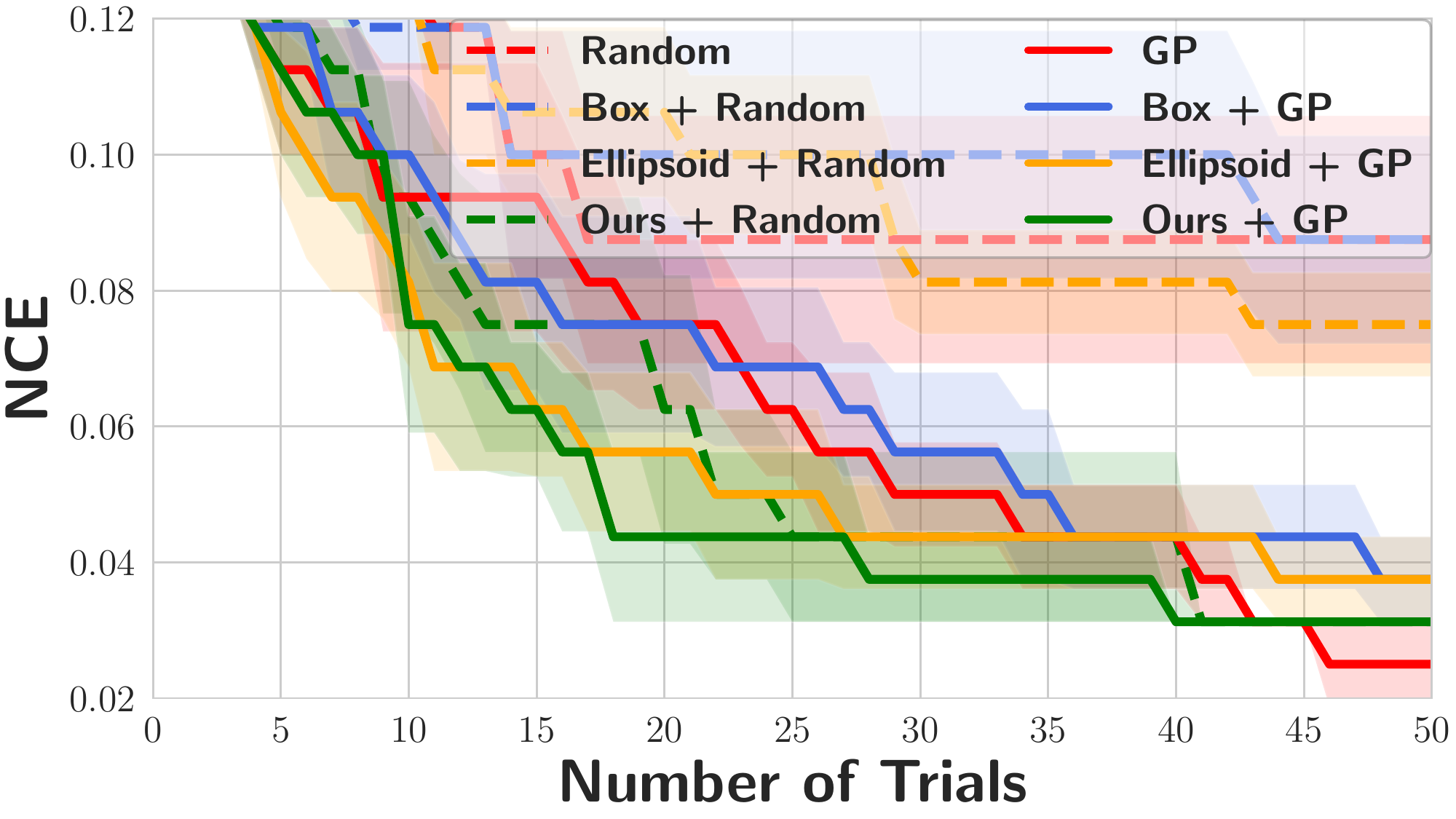}
	}}
    \subfigure[cpu\_act]{
		\scalebox{0.23}[0.23]{
			\includegraphics[width=1\linewidth]{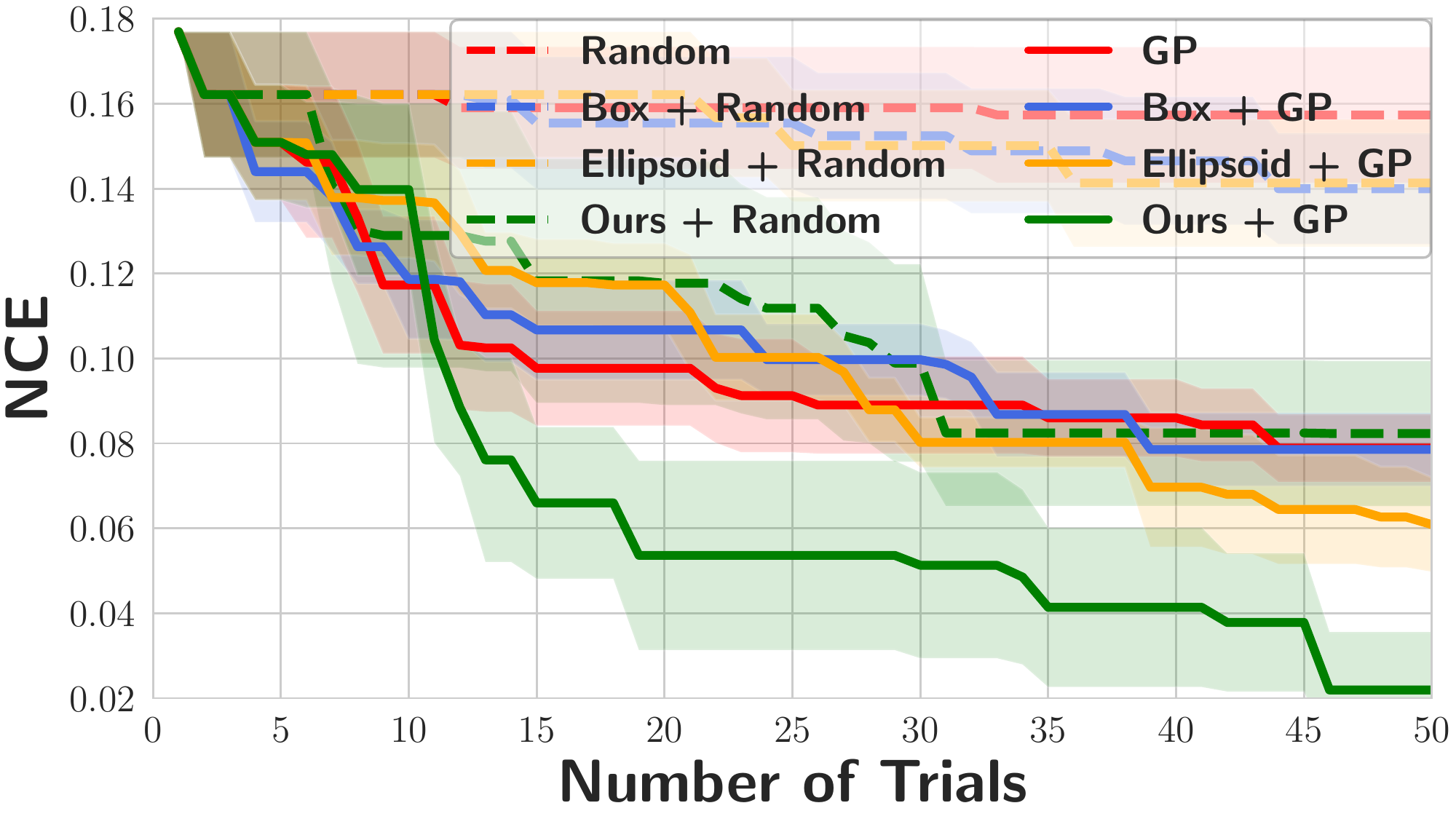}
	}}
	\vspace{-.5em}
	\caption{Normalized classification error (NCE) when tuning Random Forest on eight OpenML datasets.}
	\vspace{-.5em}
  \label{fig:exp_rf}
\end{figure*}

\begin{figure}[t!]
	\centering
	\scalebox{.65}[.65]{
	  \includegraphics[width=1\linewidth]{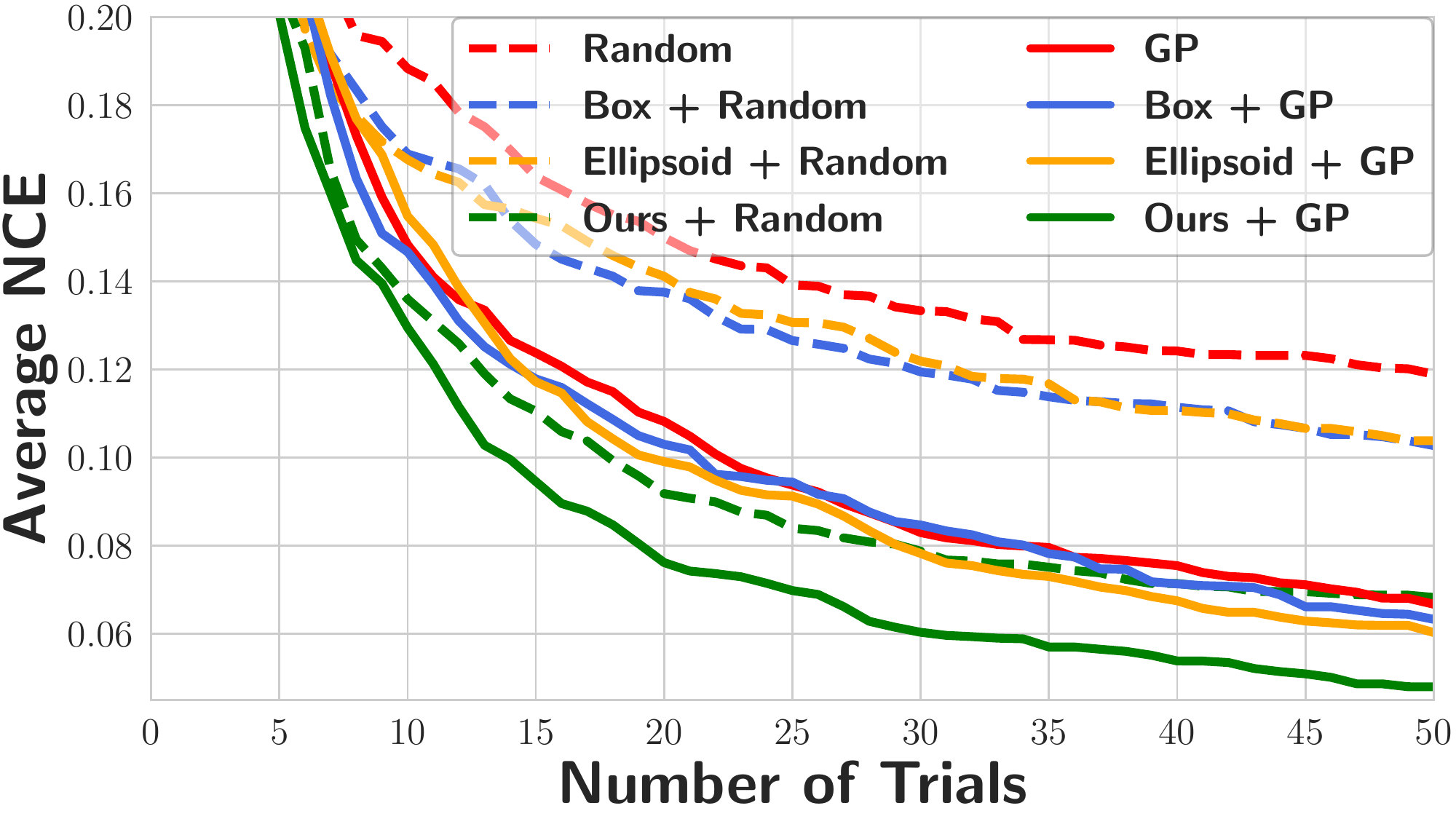}
	}
	\vspace{-.5em}
	\caption{Average NCE results when tuning random forest across 20 OpenML tasks.}
  \label{fig:exp_agg}
\end{figure}

\subsection{Experimental Settings}
\para{\textbf{Benchmarks.}}
We evaluate the performance of our proposed method on three benchmarks:
(1) Random Forest Tuning Benchmark: The benchmark tunes 5 Random Forest hyperparameters on 20 OpenML datasets~\cite{vanschoren2014openml}. Each task contains the evaluation results of 50k randomly chosen configurations.
(2) ResNet Tuning Benchmark: The benchmark tunes 5 ResNet hyperparameters on CIFAR-10, SVHN and Tiny-ImageNet. Each task contains the evaluation results of 1500 randomly chosen configurations and 500 configurations selected by running a GP-based BO.
(3) NAS-Bench-201~\cite{dong2019bench}: This is a light-weight benchmark with 6 hyperparameters for neural architecture search, which includes the statistics of 15,625 CNN models on three datasets -- CIFAR-10-valid, CIFAR-100, and ImageNet16-120.

The hyperparameter space in the Random Forest Tuning Benchmark follows the implementation in Auto-Sklearn~\cite{feurer2015efficient}, while the space in ResNet Tuning Benchmark follows the previous work~\cite{li2021mfes}.
It takes more than 50k CPU hours and 5k GPU hours to collect the first two benchmarks.
More details about the search space and datasets of the benchmarks are provided in Appendix \ref{a.1}.

\para{\textbf{Search Space Baselines.}} 
As a search space design method, we compare our proposed method with other three search space design: 
(1) No design: using the original search space without any space extraction;
(2) Box~\cite{pNIPS2019_9438}: search space design by representing the promising region as a low-volume bounding box;
(3) Ellipsoid~\cite{pNIPS2019_9438}: search space design by representing the promising region as a low-volume ellipsoid.
Based on the search space design methods, we further evaluate random search and GP-based BO~\cite{snoek2012practical} on the compact (or original) search space. 
In addition, for neural architecture search, we also evaluate the regularized evolutionary algorithm (REA)~\cite{real2019regularized} and SMAC~\cite{hutter2011sequential}, which are state-of-the-art methods on neural architecture search.
Note that, since our method enjoys high universality, it can be easily implemented into those methods.
We will further illustrate the performance of the combination of space transfer and surrogate transfer methods in Section~\ref{sec:property_study}.



\para{\textbf{Basic Settings.}}
To evaluate the performance of the considered transfer learning method, the experiments are performed in a leave-one-out fashion, i.e., each method optimizes the hyperparameters of a specific task while treating the remaining tasks as the source tasks. 
We follow the experiment settings as in previous work~~\cite{wistuba2016two, feurer2018scalable}.
Concretely, only $N_S$ observations in the source tasks (here $N_S=100$) are used to extract promising region.
In addition, following~\cite{feurer2018scalable}, all the compared methods are initialized with three randomly selected configurations, and then
a total of $N_{T}=50$ configuration evaluations are conducted sequentially to evaluate the effectiveness of the transfer learning based space design given limited trials. 
To avoid the effect of randomness, each method is repeated $20$ times and the mean performance metrics are reported along with their variance.

\para{Evaluation Metrics.}
For each task, the objective of HPO is to find the configuration with the lowest validation error.
However, since the classification error is not commensurable across the benchmark datasets, we follow the previous work~\cite{bardenet2013collaborative,wistuba2016two,feurer2018scalable} and report the Normalized Classification Error (NCE) of all compared baselines on each dataset. The NCE after $t$ trials is defined as:
\begin{equation}
NCE(\mathcal{X}^i_t) = \frac{min_{\bm{x} \in \mathcal{X}_t} y^{i}_{\bm{x}} - y^{i}_{min}}{y^i_{max} - y^{i}_{min}},
\end{equation}
where $y^i_{min}$ and $y^i_{max}$ are the best and worst ground-truth performance value (i.e., classification error) on the $i$-th task, $y_{\bm{x}}^i$ corresponds to the performance of configuration $\bm{x}$ in the $i$-th task, and $\mathcal{X}^i_t$ is the set of hyperparameter configurations on the $i$-th task that have been evaluated in the previous $t$ trials.
To measure a method on the entire benchmark, we average the NCE over all considered tasks, that is, $\frac{1}{K}\sum_{i=1}^KNCE(\mathcal{X}^i_t)$, where $K$ is the number of tasks.




\para{Implementations \& Parameters.}
Please refer to Appendix \ref{reproduction} for more details about implementations and reproductions.

\begin{figure*}[htb]
	\centering
	\subfigure[CIFAR-10]{
		\scalebox{0.27}[0.27]{
			\includegraphics[width=1\linewidth]{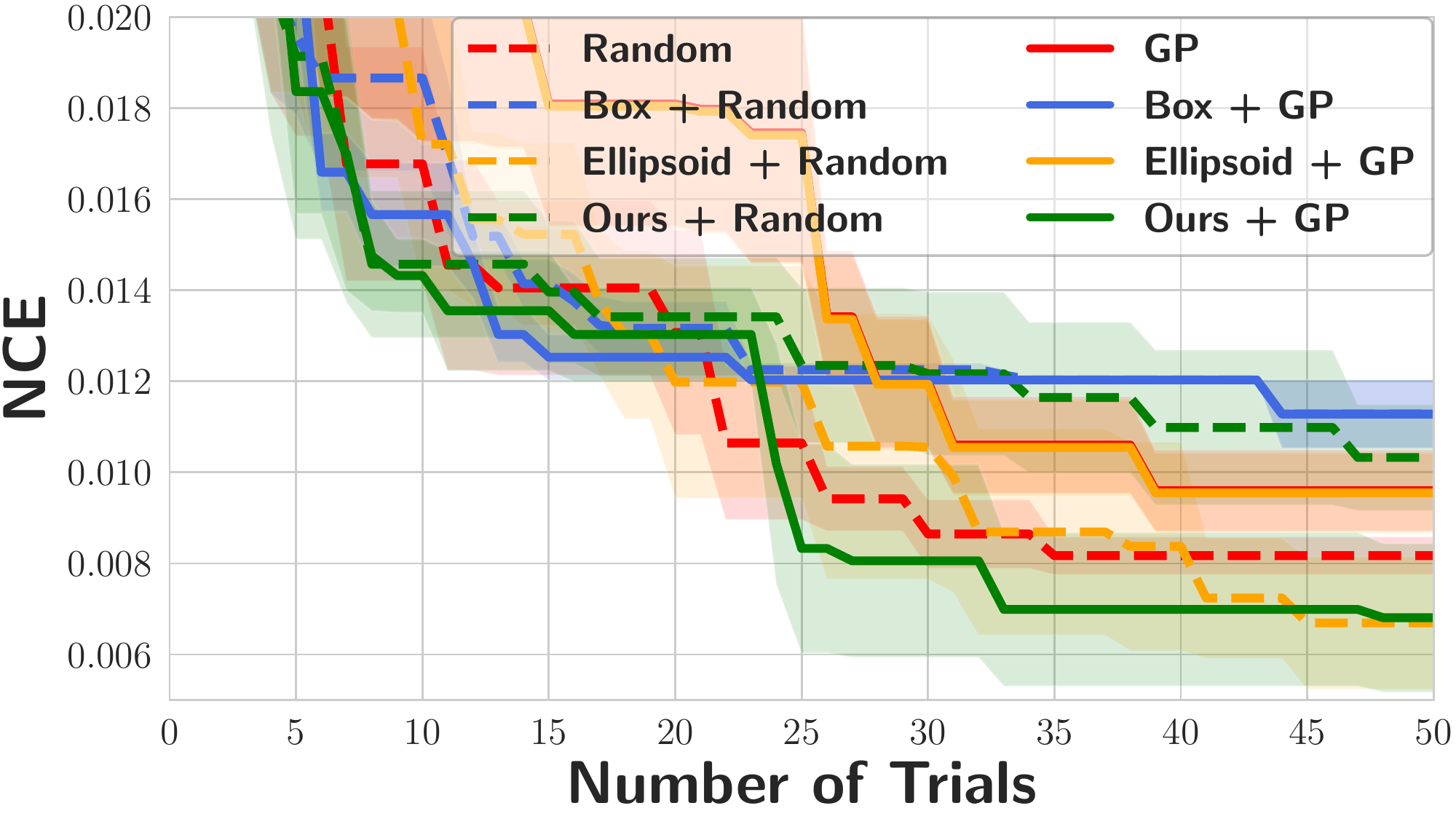}
	}}
	\subfigure[SVHN]{
		\scalebox{0.27}[0.27]{
			\includegraphics[width=1\linewidth]{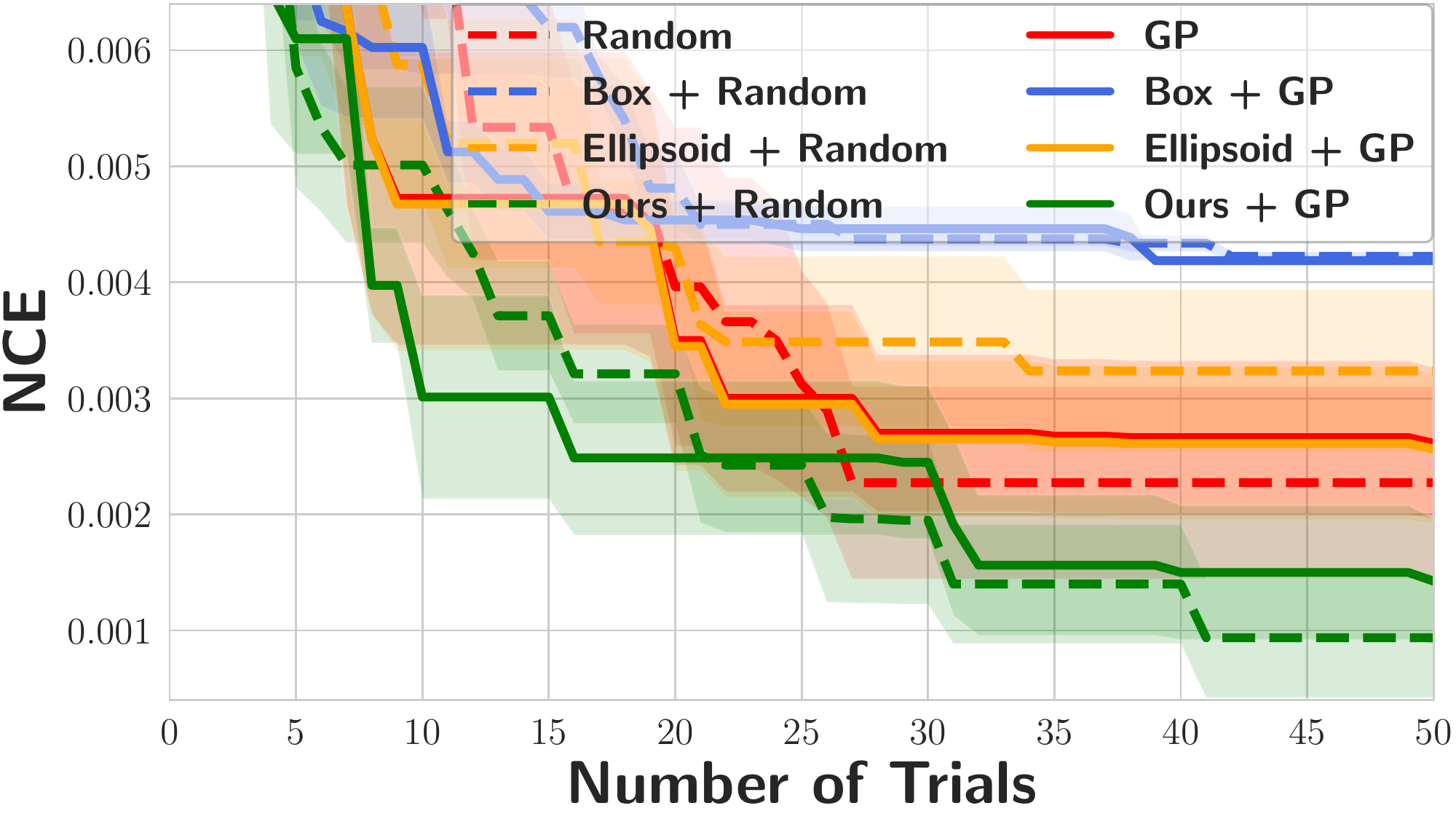}
	}}
	\subfigure[Tiny-ImageNet]{
		\scalebox{0.27}[0.27]{
			\includegraphics[width=1\linewidth]{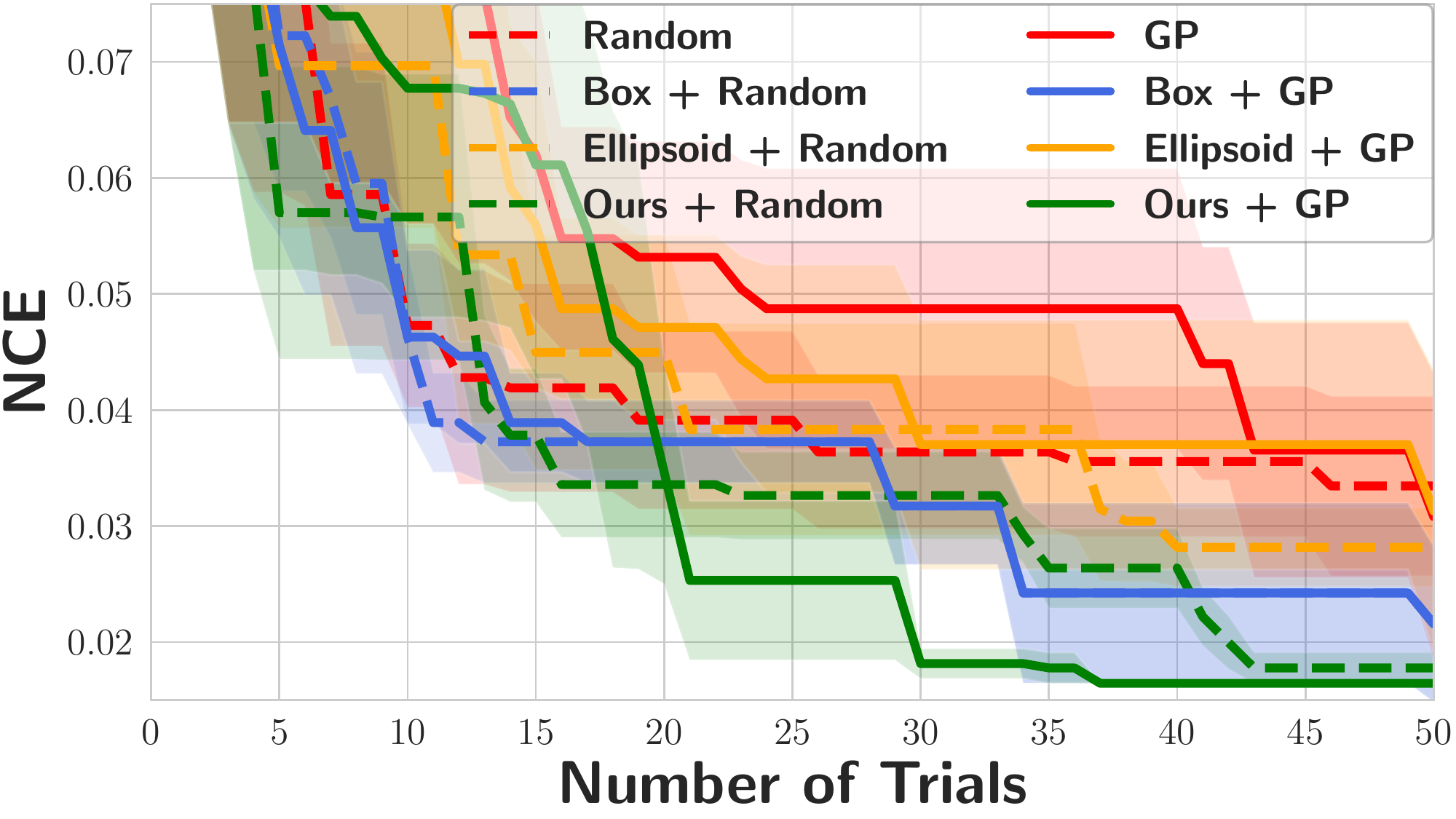}
	}}
	\vspace{-1.em}
	\caption{Normalized classification error (NCE) when tuning ResNet on three vision problems.}
	\vspace{-1.em}
  \label{fig:exp_resnet}
\end{figure*}

\begin{figure*}[htb]
	\centering
	\subfigure[CIFAR-10]{
		\scalebox{0.27}[0.27]{
			\includegraphics[width=1\linewidth]{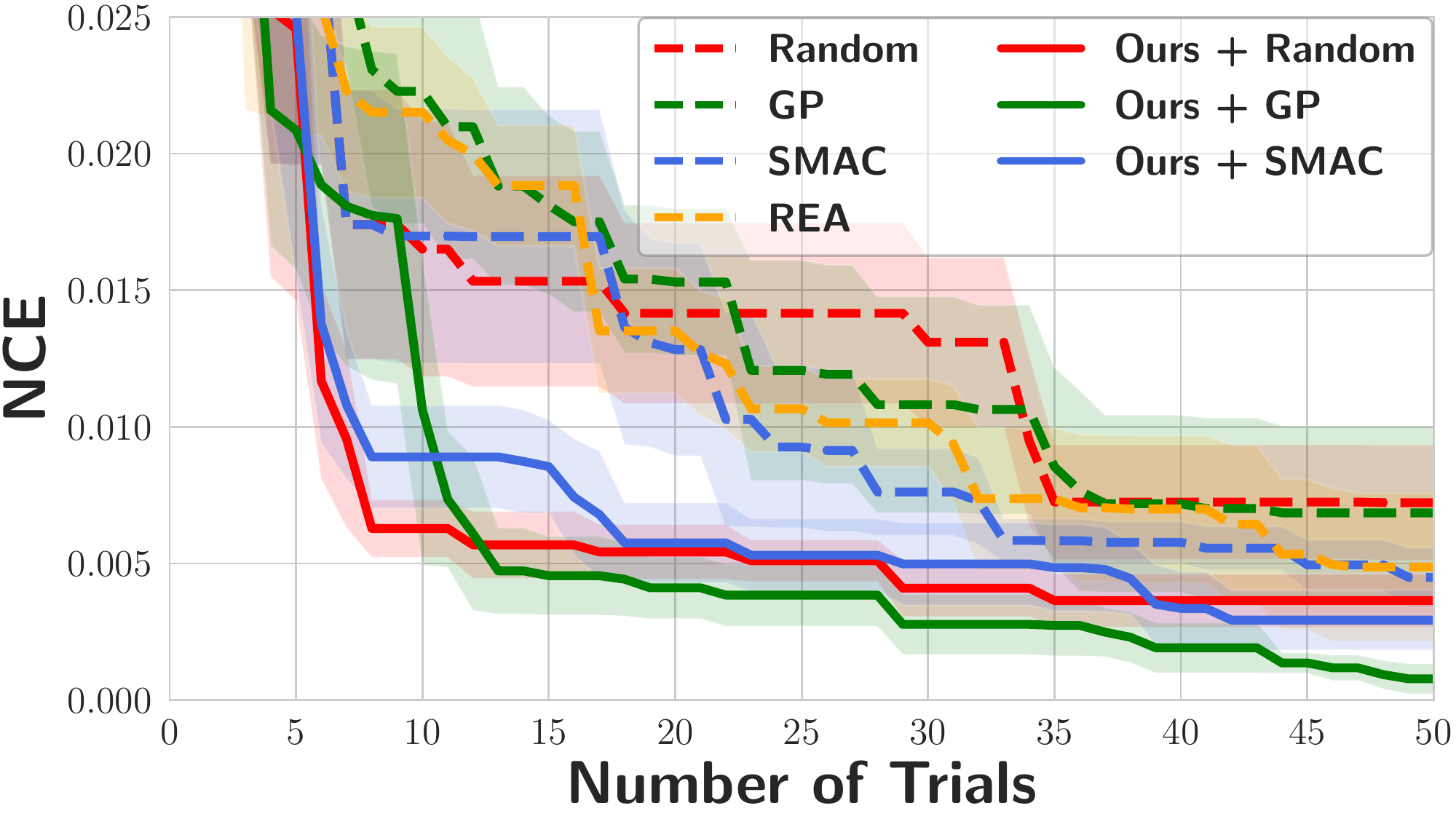}
	}}
	\subfigure[CIFAR-100]{
		\scalebox{0.27}[0.27]{
			\includegraphics[width=1\linewidth]{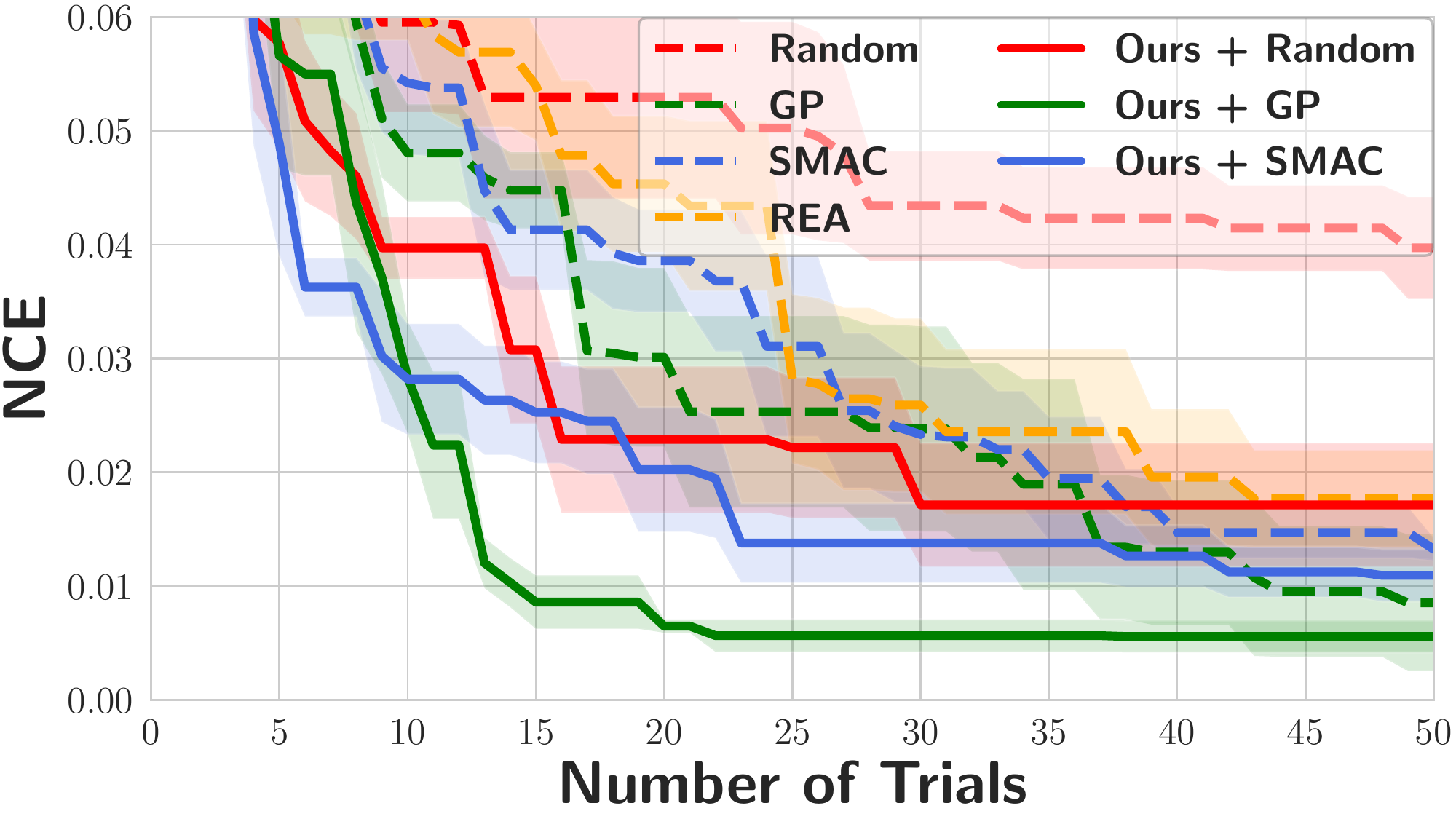}
	}}
	\subfigure[ImageNet16-120]{
		\scalebox{0.27}[0.27]{
			\includegraphics[width=1\linewidth]{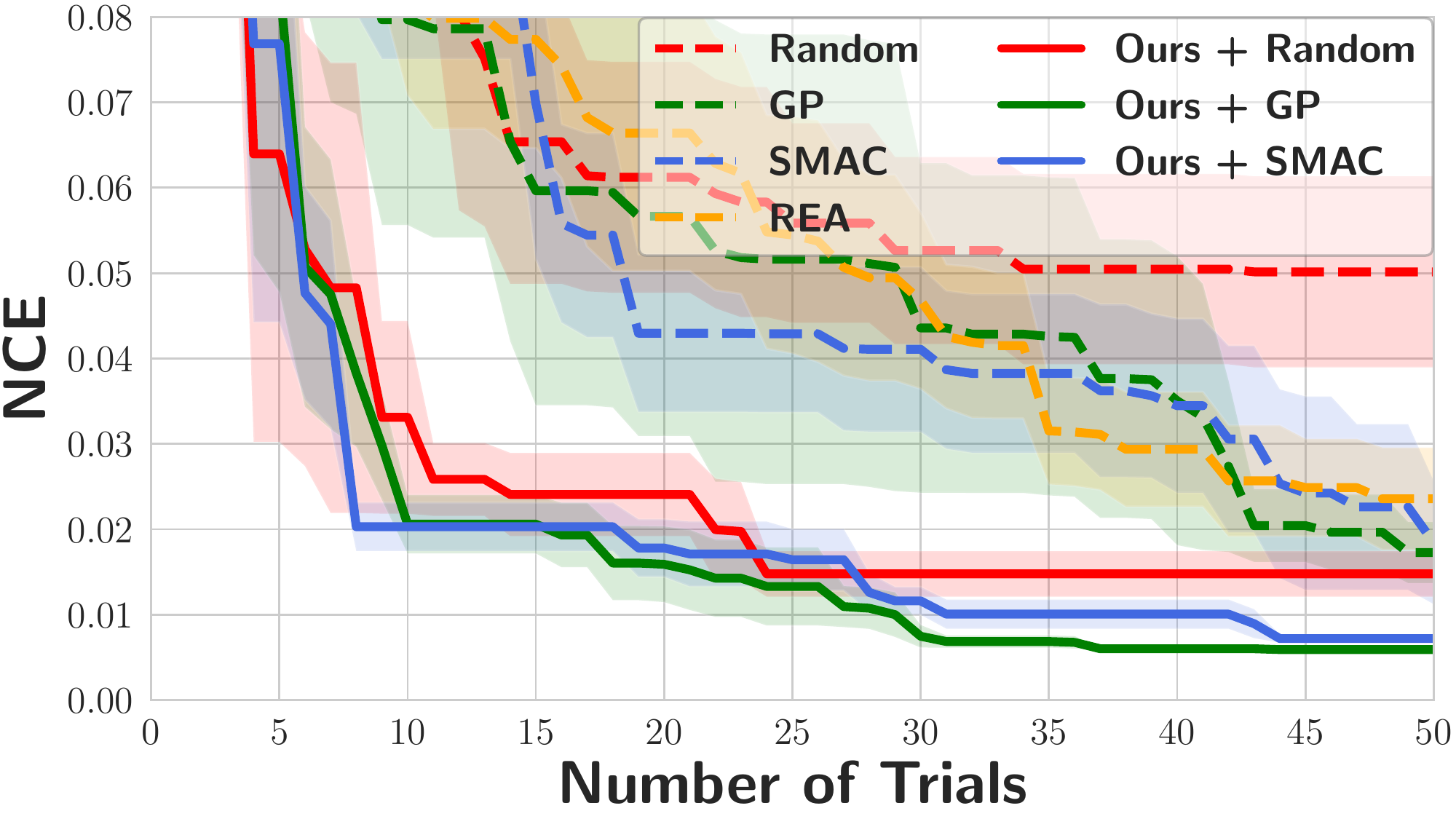}
	}}
	\vspace{-1.em}
	\caption{Normalized classification error (NCE) when conducting architecture search on NASBench201.}
	\vspace{-.5em}
  \label{fig:exp_nas}
\end{figure*}

\subsection{Tuning Random Forest on OpenML Tasks}
We first compare our method with other space design methods on the Random Forest Tuning Benchmark. 
Concretely, each task is selected as the target task in turn, and the remaining 19 tasks are the source tasks.
The search space from each space method is further evaluated by random search and GP-based BO.
Figure~\ref{fig:exp_rf} demonstrates the normalized classification error (NCE) on 8 datasets and Figure~\ref{fig:exp_agg} shows the aggregated NCE on the entire benchmark.
We observe that the Ellipsoid variants perform better than the Box and non-transfer counterparts on most tasks in terms of convergence speed. 
However, on musk and satimage, GP works slightly better than Ellipsoid GP and Box GP.
As we have explained in Section~\ref{sec:intro}, the reason is that the two space design methods ignore the relevancy between tasks, which hampers the performance of transfer learning when using dissimilar source tasks.
Generally, the variants of our method consistently outperform their counterparts.
The NCE of Ours + GP rapidly drops in the first 25 iterations, and achieves the best final performance on 7 out of 8 tasks.
In addition, we find that random search equipped with our method outperforms the Box and Ellipsoid variants on sick and musk, which further indicates the superiority of our space design method.



\subsection{Tuning ResNet on Vision Problems}
In this part, we evaluate our method on a deep learning algorithm tuning problem.
Figure~\ref{fig:exp_resnet} shows the compared results on three vision problems. 
On CIFAR-10 and SVHN, we observe that GP and Ellipsoid + GP show almost the same performance.
The reason is that, the optimal configuration of Tiny-ImageNet is quite far away from the other two tasks. 
And then, it will take a large ellipsoid to cover the optimal configurations of all source tasks, which is almost the same as the original space.
In this case, they achieve almost the same performance due to the same underlying search space.
Similar to the Random Forest Tuning Benchmark, the variants of our method consistently outperforms the compared counterparts.
Remarkably, both two variants of our method reduces the NCE of the second-best baseline Box + GP by 23.7\% on Tiny-ImageNet. 

\subsection{Architecture Search on NASBench201}
Different from the above two benchmarks, the search space for neural architecture search (NAS) contains only categorical hyperparameters.
In this case, the Box and Ellipsoid variants fail to find a compact space and revert to standard search space since there is no partial order between different values of categorical hyperparameters.
Therefore, we compare our method with non-transfer counterparts.
To further demonstrate we extend the optimization algorithms with state-of-the-art SMAC and REA and the results are shown in Figure~\ref{fig:exp_nas}. 
On all the three datasets, we observe that the variants of our method consistently performs better than non-transfer counterparts.
When we equip the state-of-the-art method SMAC with our method, we observe a significant reduction on NCE, which indicates that our method enjoys practicality on different types of HPO tasks.



\begin{figure}[t!]
\centering
	\subfigure[Universality]{
		\scalebox{0.47}[0.47]{
			\includegraphics[width=1\linewidth]{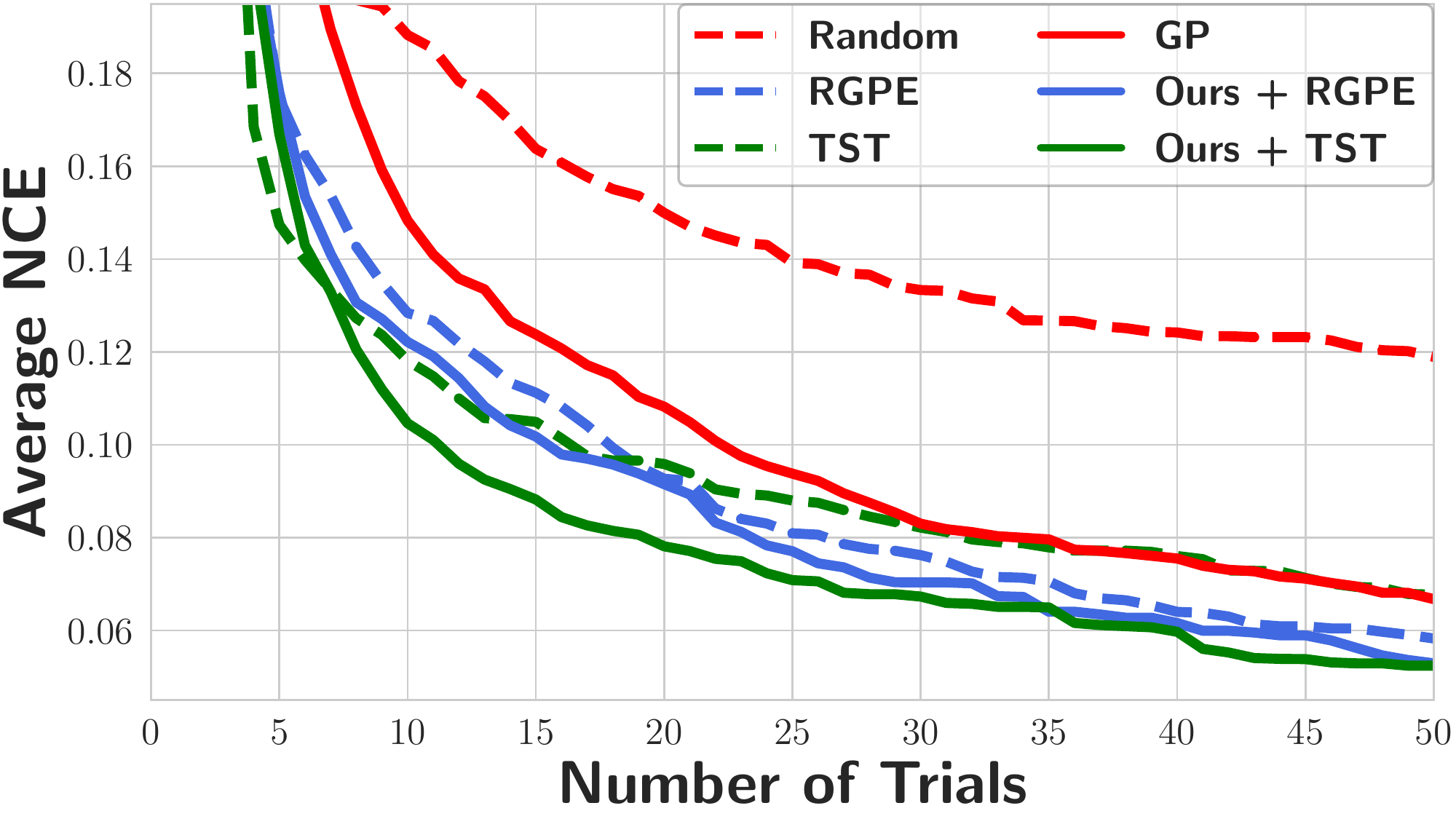}
			\label{fig:exp_rgpetst}
	}}
	\subfigure[Safeness]{
		\scalebox{0.47}[0.47]{
			\includegraphics[width=1\linewidth]{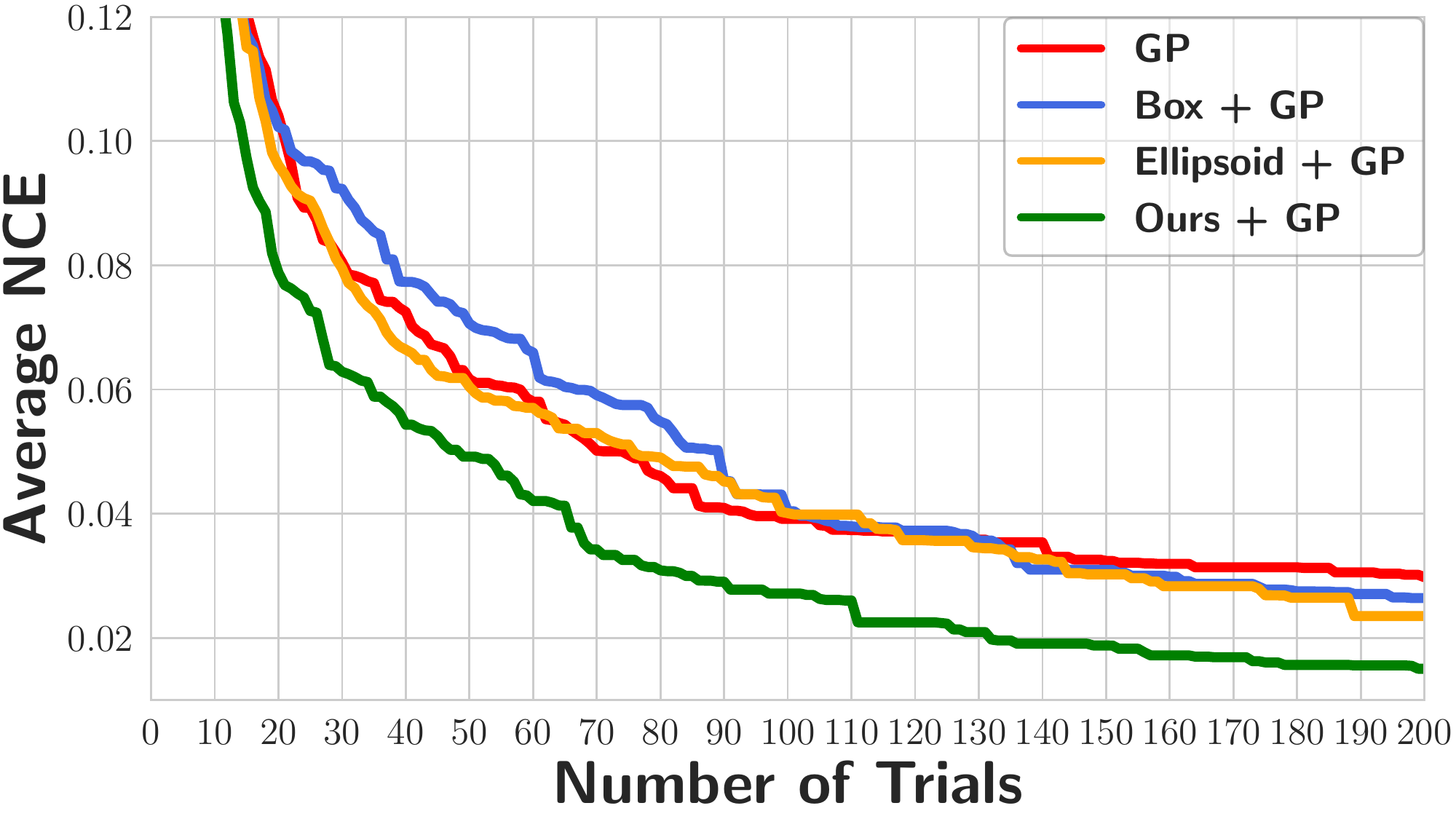}
		\label{fig:exp_long}
	}}
	\vspace{-1.em}
	\caption{Case study on universality and safeness.}
    \label{fig:exp_universal}
\end{figure}

\subsection{Case Study on Universality and Safeness}
\label{sec:property_study}
In this subsection, we investigate the advantages of our proposed method.
We first study how our method works with existing surrogate transfer methods (Universality) on Random Forest Tuning Benchmark.
We choose the state-of-the-art RGPE~\cite{feurer2018scalable} and TST~\cite{wistuba2016two} as the optimization algorithms.
Figure~\ref{fig:exp_rgpetst} demonstrates the optimization results with and without the surrogate transfer methods on Random Forest Tuning Benchmark.
We observe that the surrogate transfer methods alone (RGPE and TST) indeed accelerate hyperparameter optimization.
When our method is implemented, the performance is further improved. 
Concretely, our method reduces the aggregated NCE of
RGPE and TST by 10.1\% and 22.6\% on the entire benchmark, respectively.

In addition, we set the number of trials to 200 showcase the convergence results of compared methods given a larger budget (Safeness). 
Figure~\ref{fig:exp_long} shows that our method still outperforms the compared GP variants.
Concretely, our method with GP reduces the aggregated NCE by 36.0\% compared with the second-best baseline Ellipsoid + GP on the Random Forest Tuning Benchmark.

\begin{figure}[t!]
	\centering
	\subfigure[musk]{
		\scalebox{0.4}[0.4]{
			\includegraphics[width=1\linewidth]{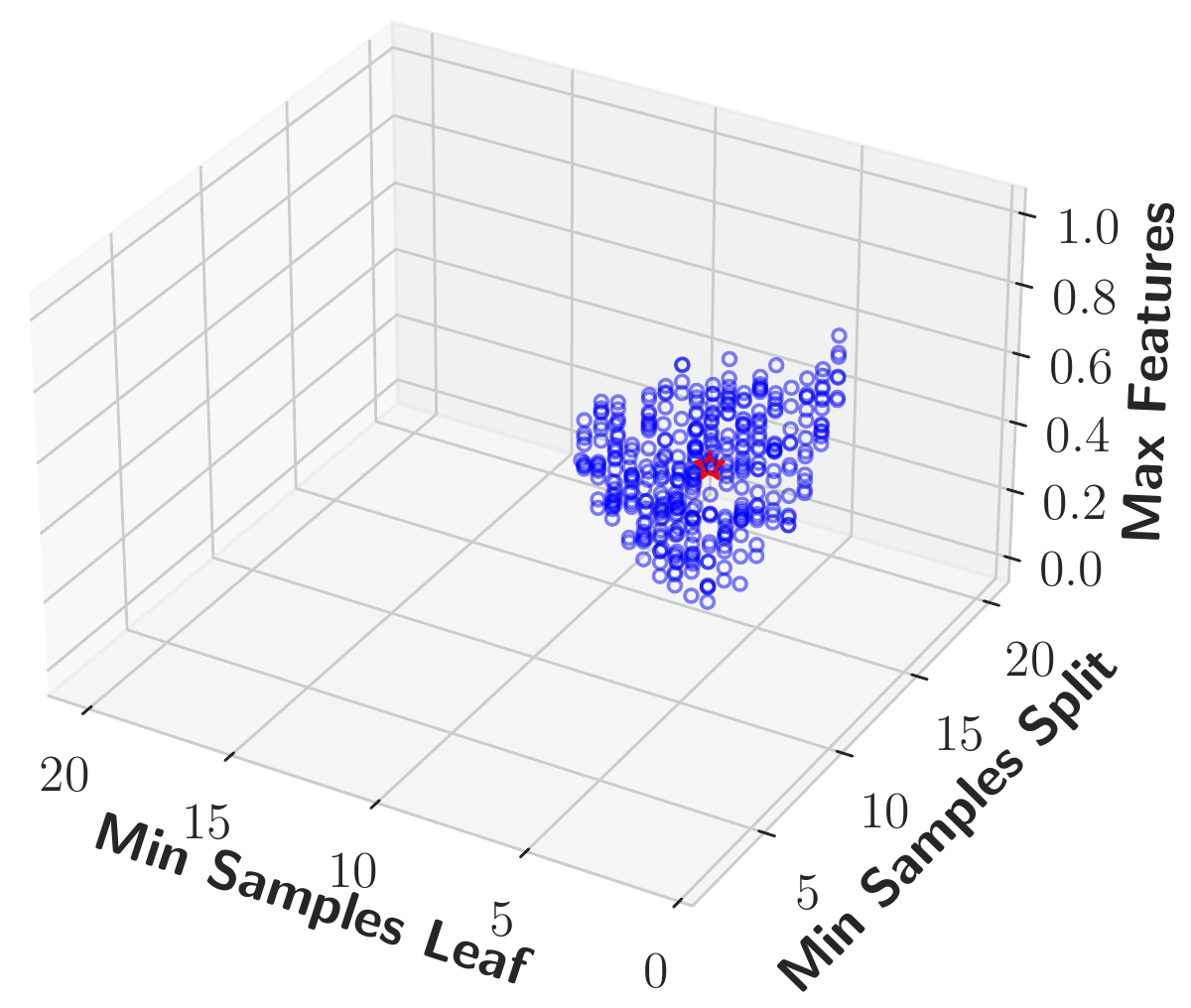}
	}}
	\subfigure[cpu\_act]{
		\scalebox{0.4}[0.4]{
			\includegraphics[width=1\linewidth]{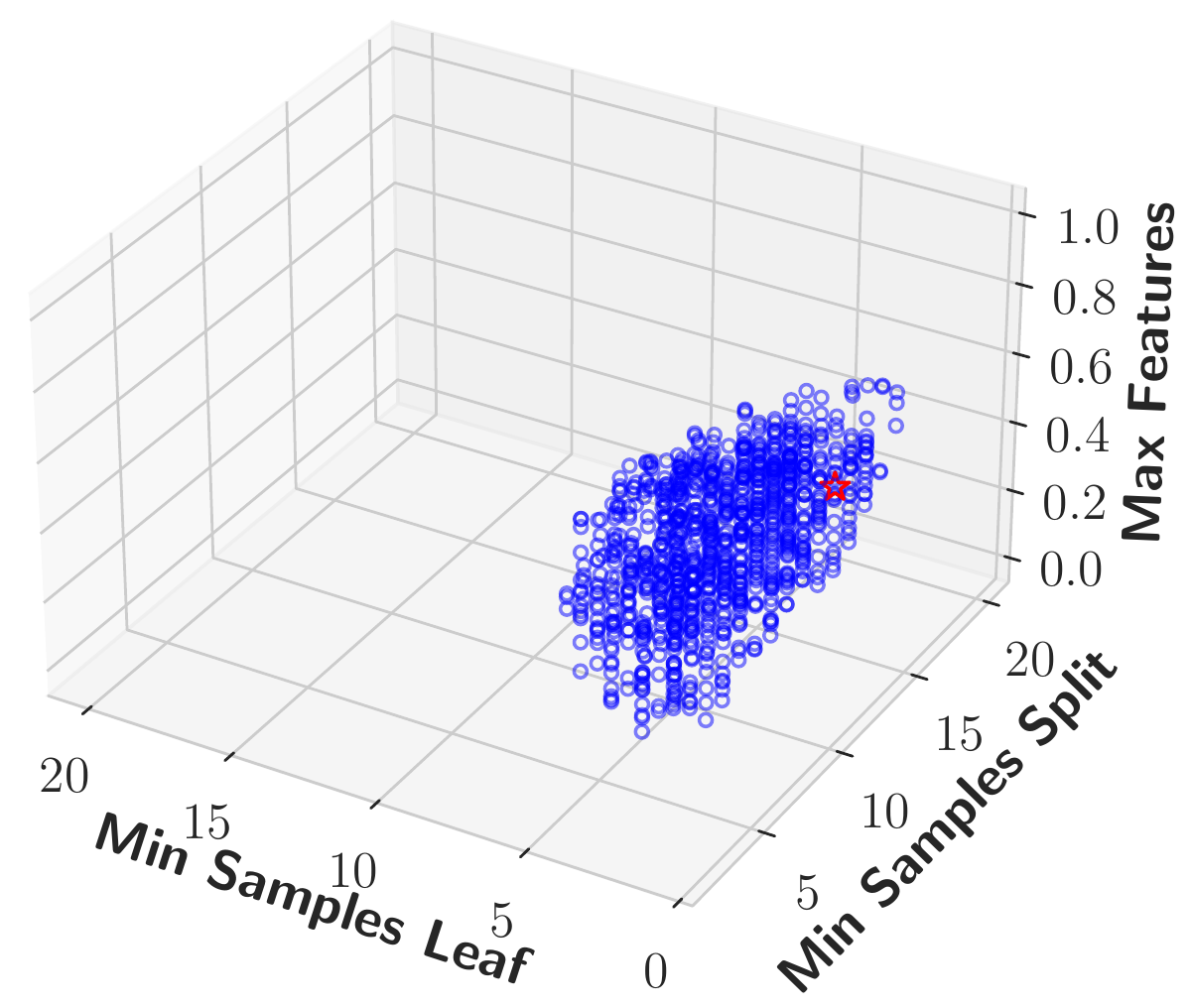}
	}}
	\caption{Candidate points in our compact search space. The blue points refer to candidates and the red points refer to the global optimum.}
  \label{fig:exp_visual_real}
\end{figure}

\subsection{Case Study on Search Space Design}
In this subsection, we visualize the compact search space given by our method on two datasets in Random Forest Tuning Benchmark.
Figure~\ref{fig:exp_visual_real} plots the candidate points in our compact search space during the 50-th iteration on two datasets. 
Rather than optimizing over the entire cube, our method generates a significantly smaller search space, which still contains the global optimum (the red point in Figure~\ref{fig:exp_visual_real}).
Remarkably, the size of search space designed by our method is 375 and 1904 on musk and cpu\_act, respectively. 
Compared with the original search space with 50000 candidates, the size of space shrinks to 0.75\% and 3.81\% of the original search space on the two tasks.

\begin{figure}[t!]
	\centering
		\scalebox{1.0}[1.0] {
		\includegraphics[width=1\linewidth]{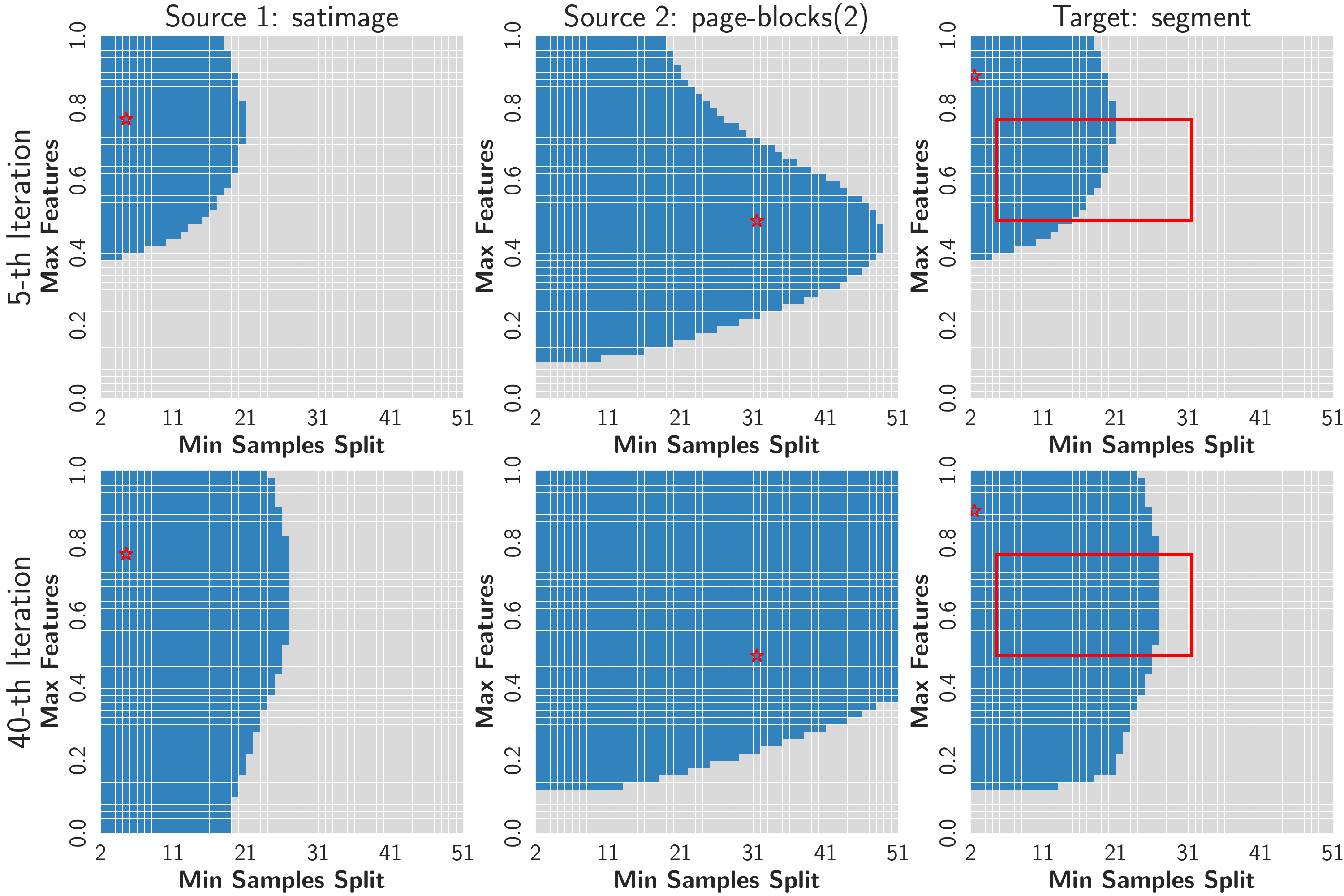}
         }
	\caption{Promising regions for source tasks and target search space when tuning on segment. The blue area refers to the promising regions of the source tasks and the search space of the target task from our method. The red point is the ground-truth optimum on that task, and the red box is the search space generated by the baseline Box.}
    \label{fig:ablation_vis_segment}
\end{figure}

\begin{figure}[t!]
	\centering
		\scalebox{1.0}[1.0] {
		\includegraphics[width=1\linewidth]{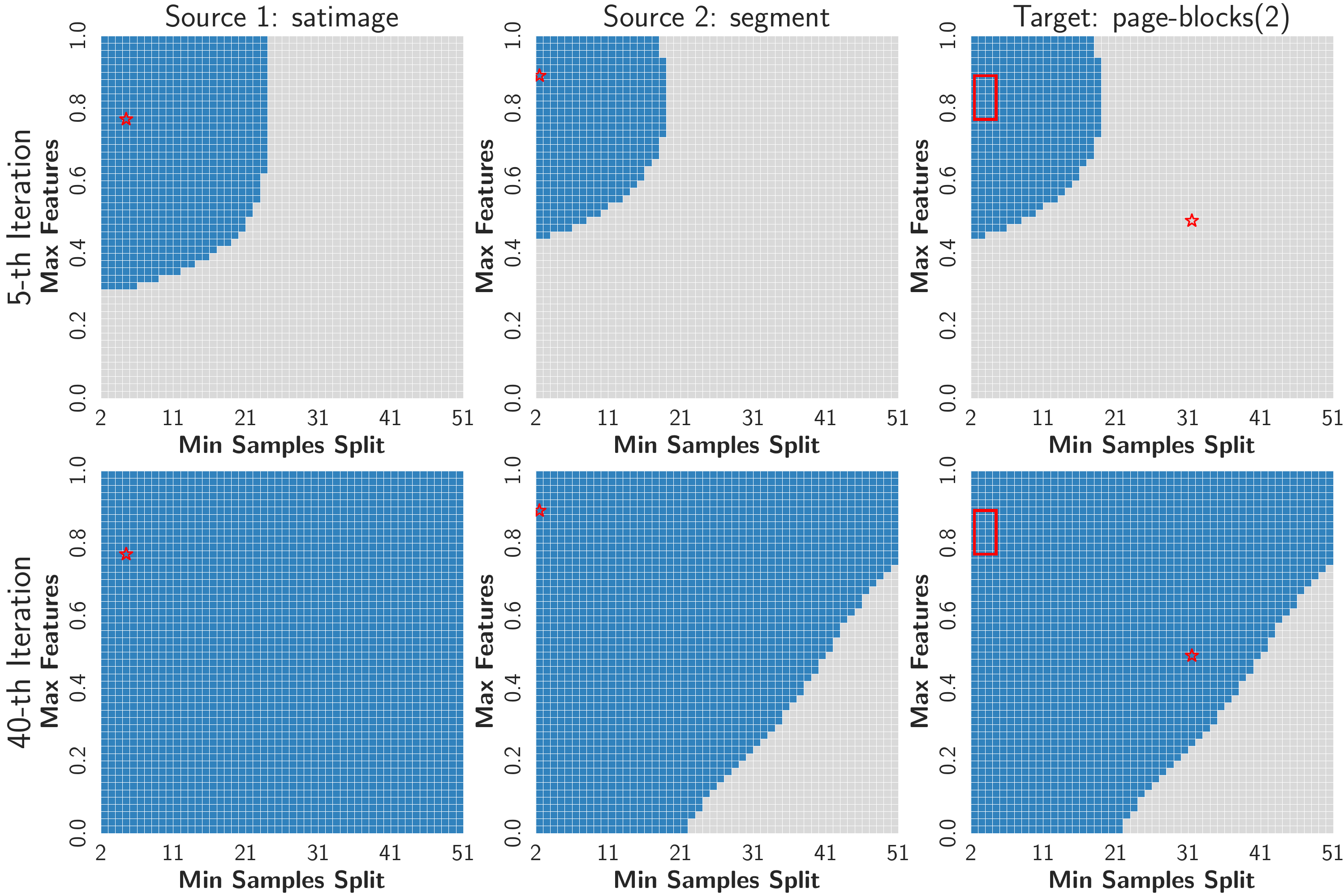}
         }
	\caption{Promising regions for source tasks and target search space when tuning on page-blocks(2).}
    \label{fig:ablation_vis_pageblocks}
\end{figure}

In addition, we also plot the promising regions (source tasks) and the compact search space (target task) in Figures~\ref{fig:ablation_vis_segment} and~\ref{fig:ablation_vis_pageblocks} using three datasets from Figure~\ref{fig:intro_heatmap}.
Recall that satimage is similar to segment, but different from page-blocks(2).
The red bounding box refers to the search space from the baseline Box.
As it only depends on the best configurations found in the source tasks, we observe that neither of the two search spaces from Box covers the global optimum (the red point in Figures~\ref{fig:ablation_vis_segment} and~\ref{fig:ablation_vis_pageblocks}) on the target task.
In Figure~\ref{fig:ablation_vis_segment} where the first source task is similar to target task, our compact search space covers the global optimum since the 5-th iteration.
However, as shown in Figure~\ref{fig:ablation_vis_pageblocks} where the target task is quite dissimilar to both source tasks, the compact search space does not include the global optimum in the very beginning. 
In this case, the promising region of source tasks gradually expand due to the dynamic quantile introduced in Section~\ref{sec:prg}, and it eventually covers the global optimum in the 40-th iteration while still excluding the bad regions as shown in Figure~\ref{fig:intro_heatmap}.
This demonstrates the effectiveness and safeness of our approach.

\begin{figure}[t!]
\centering
	\subfigure[Promising Region Extraction]{
		\scalebox{0.47}[0.47]{
			\includegraphics[width=1\linewidth]{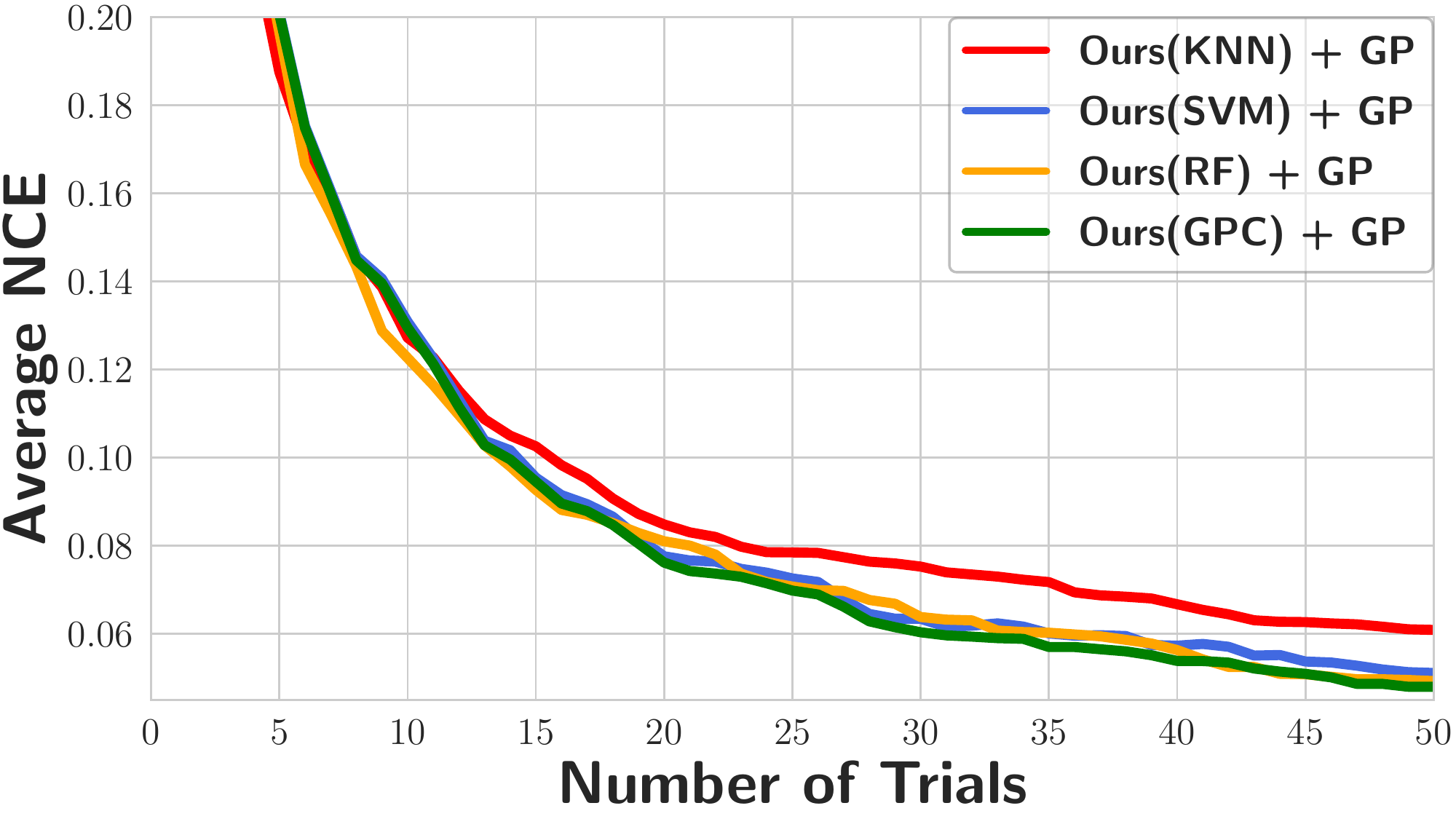}
			\label{fig:ab_c}
	}}
	\subfigure[Target Search Space Generation]{
		\scalebox{0.47}[0.47]{
			\includegraphics[width=1\linewidth]{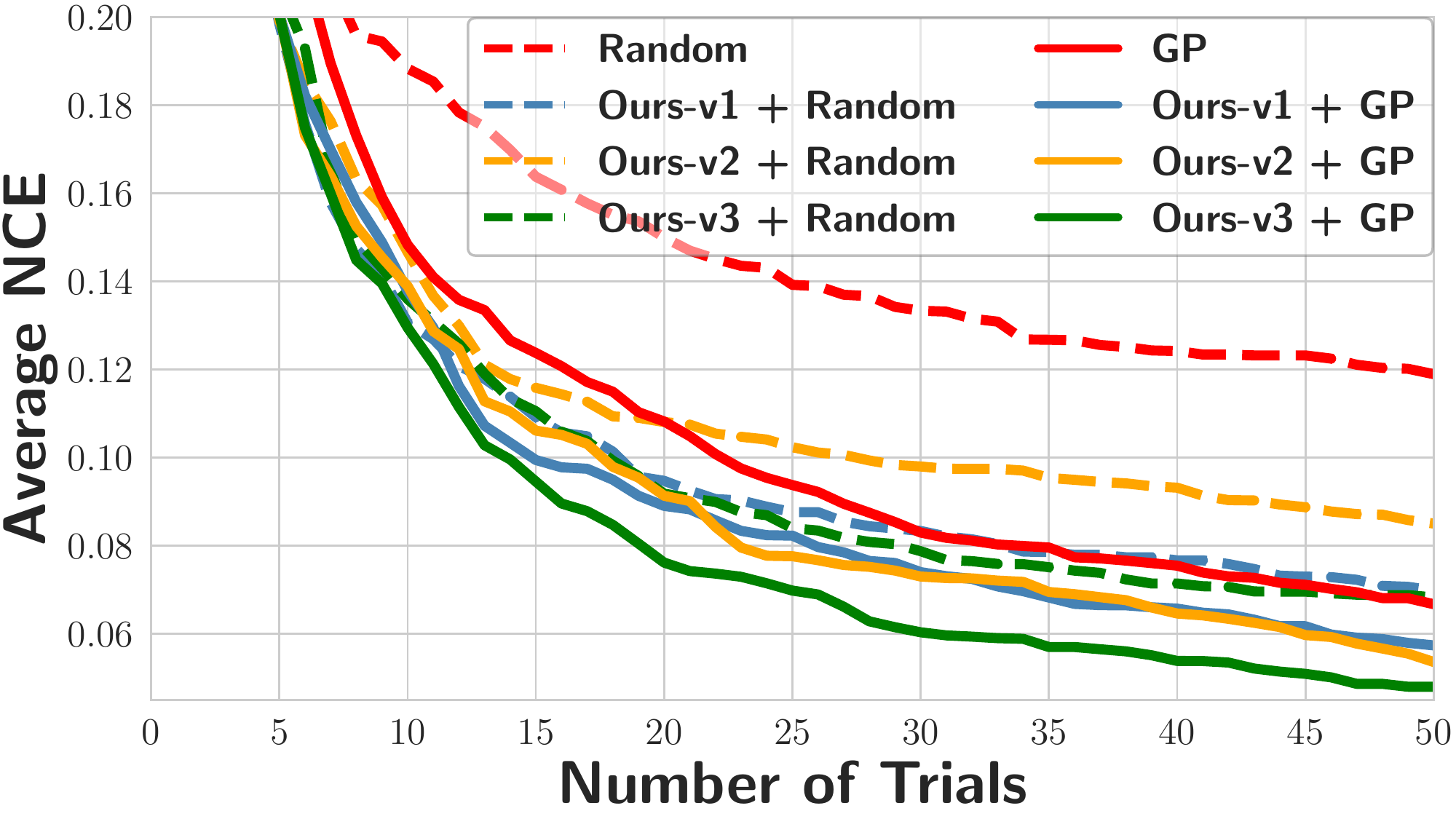}
			\label{fig:ab_e}
	}}
	\caption{Ablation study on promising region extraction and target search space generation.}
    \label{fig:ablation_method}
\end{figure}

\subsection{Ablation Study}

Finally, we provide ablation study on the machine learning classifiers used in promising region extraction and the target search space generation strategy on Random Forset Tuning Benchmark.
For classifiers, we compare the influence of different machine learning algorithms used in promising region extraction.
Figure~\ref{fig:ab_c} shows the results of four algorithm choices: KNN, LibSVM, Random Forest (RF), and Gaussian Process Classifier (GPC) used in our method.
Among the choices, the GPC shows the most stable performance across all tasks.
Moreover, GPC is a the only choice without algorithm hyperparameters, and thus we employ it as the classifier for promising region extraction.

In addition, we denote 1) OURS-v1 as using the promising region of the most similar source task; 2) OURS-v2 as using the promising region of a sampled source task; 3) OURS-v3 as our method.
Figure~\ref{fig:ab_e} demonstrates the results of those three variants. 
Among the three methods, OURS-v3 performs the best while OURS-v1 and OURS-v2 perform worse that it.
As we have explained in Section~\ref{sec:prc}, the reason is that OURS-v1 may be trapped in a sub-optimal region provided by the most similar task, and OURS-v2 can not leverage the information from various source tasks.

\section{Conclusion}

We presented a novel approach to incorporate transfer learning in BO. Rather than designing
a specialized multi-task surrogate model, our method automatically crafts promising search spaces based on previously tuning tasks. 
The extensive experiments on a wide range of tuning tasks demonstrate that our
approach could significantly speed up the HPO process and enjoy desirable properties.

\begin{acks}
This work was supported by the National Natural Science Foundation of China (No.61832001), Beijing Academy of Artificial Intelligence (BAAI), PKU-Tencent Joint Research Lab. Bin Cui is the corresponding author.
\end{acks}


\bibliographystyle{ACM-Reference-Format}
\bibliography{reference}

\clearpage

\appendix
\section{Appendix}

\subsection{The Details of Benchmark}
\label{a.1}
As described in Section 5, we create two benchmarks to evaluate the performance of search space design methods, namely the Random Forest Tuning Benchmark and ResNet Tuning Benchmark. In addition, we apply NASBench-201~\cite{dong2019bench} to test the practicality of our method on different types of hyperparameters.

\textbf{Random Forest Tuning Benchmark.}
The random forest classifier is a widely used tree-based ensemble model in data analysis.
The implementation of random forest and the design of their hyperparameter space in the benchmark follows the widely-used AutoML system --- Auto-sklearn~\cite{feurer2015efficient}. 
The range and the default value of each hyperparameter are illustrated in Tables \ref{hp_trees}. 
To collect sufficient source HPO data for transfer learning, we select 20 real-world datasets from OpenML repository~\cite{vanschoren2014openml}, and evaluate the validation and test performance (i.e., the balanced accuracy) of 50k configurations for each dataset, which are randomly sampled from the hyperparameter space. 
The datasets used in our benchmarks are of medium size, whose number of rows ranges from 2000 to 8192. 
For more details, see Table~\ref{table_dataset}. 
The total number of model evaluations (observations) in our benchmarks reaches 10 million, and it takes more than 50k CPU hours to obtain the evaluation results. 
For reproduction purposes, we also upload the benchmark data (e.g., evaluation results and the corresponding scripts) along with this submission.
The benchmark data (with size – 355.4Mb); due to the space limit (maximum 20Mb) on CMT3, we only upload a small subset of this benchmark. 
After the review process, we will make the complete benchmark publicly available (e.g., on Google Drive).

\begin{table}[h]
\centering
\small
\begin{tabular}{lccc}
    \toprule
    Hyperparameter & Range  & Default \\
    \midrule
    criterion &  \{gini, entropy\} & gini \\
    max\_features & [0, 1] & 0.5 \\
    min\_sample\_split & [2, 20] & 2 \\
    min\_sample\_leaf & [1, 20] & 1 \\
    bootstrap & \{True, False\} & True \\
    \bottomrule
\end{tabular}
\caption {Hyperparameters of Random Forest.}
\label{hp_trees}
\vskip -0.1in
\end{table}

\begin{table}[htb]
\centering
\resizebox{1\columnwidth}{!}{
\begin{tabular}{lccccc}
    \toprule
    Datasets & OpenML ID & Classes & Samples & Continuous & Nominal \\ 
    \midrule
    kc1 & 1067 & 2 & 2109 & 21 & 0 \\
    quake & 772 & 2 & 2178 & 3 & 0 \\
    segment & 36 & 7 & 2310 & 19 & 0 \\
    madelon & 1485 & 2 & 2600 & 500 & 0 \\
    space\_ga & 737 & 2 & 3107 & 6 & 0 \\
    sick & 38 & 2 & 3772 & 7 & 22 \\
    pollen & 871 & 2 & 3848 & 5 & 0 \\
    abalone & 183 & 26 & 4177 & 7 & 1 \\
    winequality\_white & - & 7 & 4898 & 11 & 0 \\
    waveform(1) & 979 & 3 & 5000 & 40 & 0 \\
    waveform(2) & 979 & 2 & 5000 & 40 & 0 \\
    page-blocks(2) & 1021 & 2 & 5473 & 10 & 0 \\
    optdigits & 28 & 10 & 5610 & 64 & 0 \\
    satimage & 182 & 6 & 6430 & 36 & 0 \\
    wind & 847 & 2 & 6574 & 14 & 0 \\
    musk & 1116 & 2 & 6598 & 167 & 0 \\
    delta\_ailerons & 803 & 2 & 7129 & 5 & 0 \\
    puma8NH & 816 & 2 & 8192 & 8 & 0 \\
    cpu\_act & 761 & 2 & 8192 & 21 & 0 \\
    cpu\_small & 735 & 2 & 8192 & 12 & 0 \\
    \bottomrule
\end{tabular}
}
\caption{Datasets used in Random Forest Tuning Benchmark.}
\vspace{-1em}
\label{table_dataset}
\end{table}

\textbf{ResNet Tuning Benchmark.}
While deep learning has attracted great attention in recent years, we also create a benchmark for DL algorithm tuning.
Concretely, we tune the ResNet with five hyperparameters as shown in Table~\ref{hp_resnet}.
The benchmark contains three vision tasks, including CIFAR-10, SVHN, and Tiny-ImageNet. 
Each task contains the evaluation results of 1500 randomly chosen configurations and 500 configurations selected by running a GP-based BO.
It takes more than 5k GPU hours to obtain the evaluation results.

\begin{table}[h]
\centering
\small
\begin{tabular}{lccc}
    \toprule
    Hyperparameter & Range  & Default \\
    \midrule
    batch size &  [32, 256] & 64 \\
    learning rate & [1e-3, 0.3] & 0.1 \\
    weight decay & [1e-5, 1e-2] & 2e-4 \\
    momentum & [0.5, 0.99] & 0.9 \\
    nesterov & \{True, False\} & True \\
    \bottomrule
\end{tabular}
\caption {Hyperparameters of ResNet.}
\label{hp_resnet}
\vskip -0.1in
\end{table}

\textbf{NASBench-201}
NASBench-201~\cite{dong2019bench} is a benchmark for neural architecture search (NAS). 
The architecture space includes six operations, and the choices for each operation are provided in Table~\ref{hp_nas}.
To help compare different NAS algorithms, NASBench-201 exhaustively evaluates all the configurations from its search space on three datasets (CIFAR-10, CIFAR-100, ImageNet16-120).
Each task includes the results of 15625 evaluations, including the validation accuracy, test accuracy, and training time.

\begin{table}[h]
\centering
\small
\resizebox{1\columnwidth}{!}{
\begin{tabular}{lccc}
    \toprule
    Hyperparameter & Choice & Default \\
    \midrule
    Operation 1 &  \{None, Skip, 1*1 Conv, 3*3 Conv, 3*3 Pooling\} & Skip \\
    Operation 2 &  \{None, Skip, 1*1 Conv, 3*3 Conv, 3*3 Pooling\} & Skip \\
    Operation 3 &  \{None, Skip, 1*1 Conv, 3*3 Conv, 3*3 Pooling\} & Skip \\
    Operation 4 &  \{None, Skip, 1*1 Conv, 3*3 Conv, 3*3 Pooling\} & Skip \\
    Operation 5 &  \{None, Skip, 1*1 Conv, 3*3 Conv, 3*3 Pooling\} & Skip \\
    Operation 6 &  \{None, Skip, 1*1 Conv, 3*3 Conv, 3*3 Pooling\} & Skip \\
    \bottomrule
\end{tabular}
}
\caption {Hyperparameters of NASBench-201.}
\label{hp_nas}
\vskip -0.1in
\end{table}

\subsection{Implementation and Reproduction Details}
\label{reproduction}

We use the implementation of Bayesian optimization in SMAC~\cite{hutter2011sequential,Lindauer2021SMAC3AV}, a toolkit for black-box optimization that support a complex hyperparameter space, including numerical, categorical, and conditional hyperparameters. 
The kernel hyperparameters in Gaussian Process are inferred by maximizing the marginal likelihood. 
In the BO module, the popular EI acquisition function is used.
The other baselines are implemented following their original papers. 
In all experiments, the $\alpha_{max}$ and $\alpha_{min}$ in our approach are set to 0.95 and 0.05, respectively.
The sampling framework samples $k=5$ tasks during each iteration on the benchmarks.

The source code and benchmark data are uploaded along with this submission on CMT3, and the source code is also available in the anonymous repository~\footnote{https://anonymous.4open.science/r/2022-5263-XXXXX} now.
To reproduce the experimental results in this paper, an environment of Python 3.7+ is required. We introduce the experiment scripts and installation of required tools in \emph{README.md} and list the required Python packages in \emph{requirements.txt} under the root directory. 
Take one experiment as an example, to evaluate the offline performance of our method with random search and other baselines on Random Forest Tuning Benchmark with 50 trials, just execute the following script: \newline
{\em
python tools/offline\_benchmark.py --algo\_id random\_forest --trial\_num 50 --methods rs,box-rs,ellipsoid-rs,ours-rs --rep 20
}

Please check the document \emph{README.md} in this repository for more details, e.g., how to use the benchmark data in this benchmark, and how to run the experiments.

\end{document}